\documentclass[3p]{elsarticle}
\pdfoutput=1
\usepackage[utf8]{inputenc}
\usepackage{graphicx}
\usepackage{subcaption}  % Keep only this, remove subfig
\usepackage{hyperref, amsmath, amsfonts, natbib, bm, floatrow, multirow, wrapfig}
\usepackage[dvipsnames]{xcolor}
\usepackage{ulem}

\graphicspath{{Figures/}}

\journal{Journal name}

\begin{document}

\begin{frontmatter}
  \begin{center}  
  { \fontsize{16}{10} \textbf{UP-dROM : Uncertainty-Aware and Parametrised dynamic Reduced-Order Model -- application to unsteady flows} } \\
    \vspace{0.5cm}
        \begin{figure}[ht]
        \centering
        \includegraphics[width=0.17\textwidth]{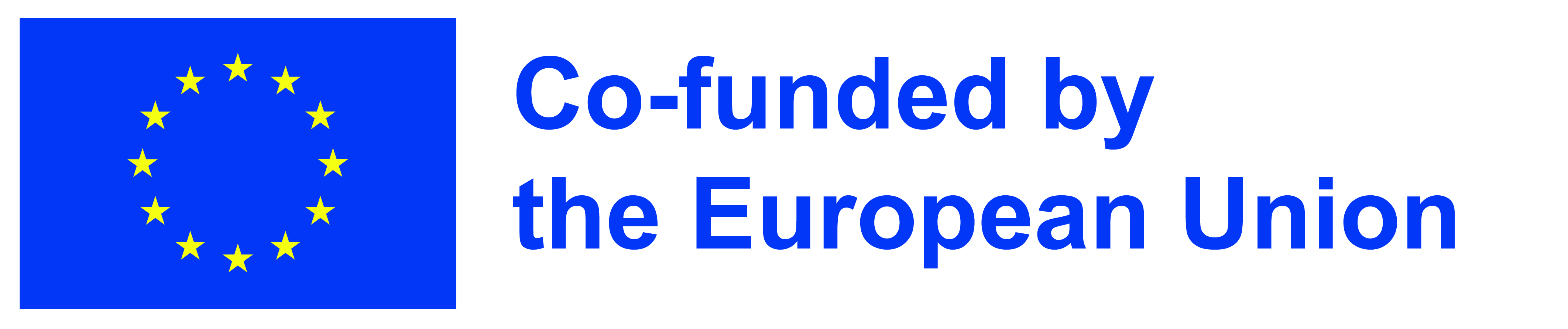}\hspace{1cm}
        \includegraphics[width=0.17\textwidth]{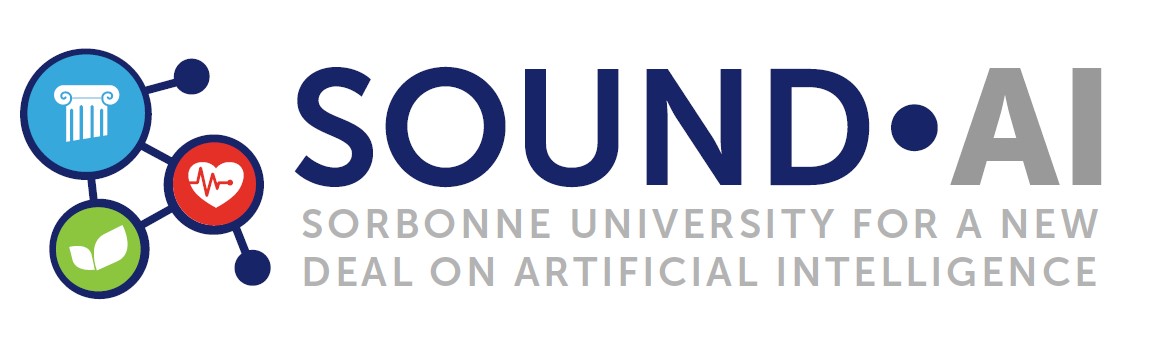}\hspace{1cm}
        \includegraphics[width=0.12\textwidth]{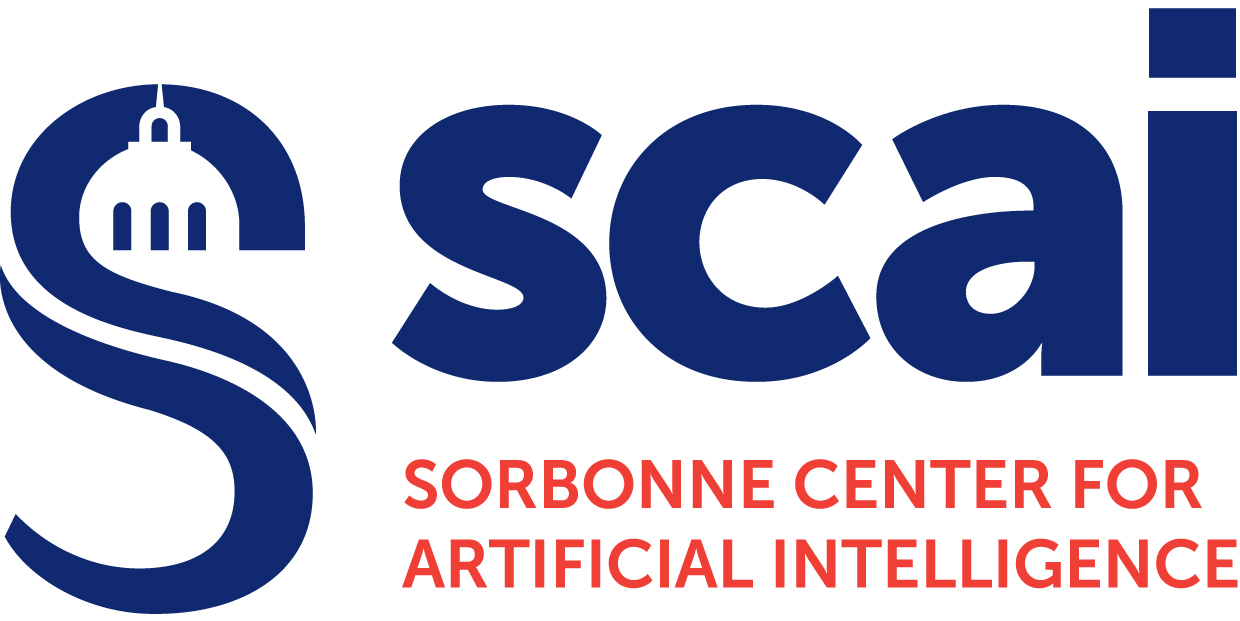}\hspace{1cm}
        \includegraphics[width=0.12\textwidth]{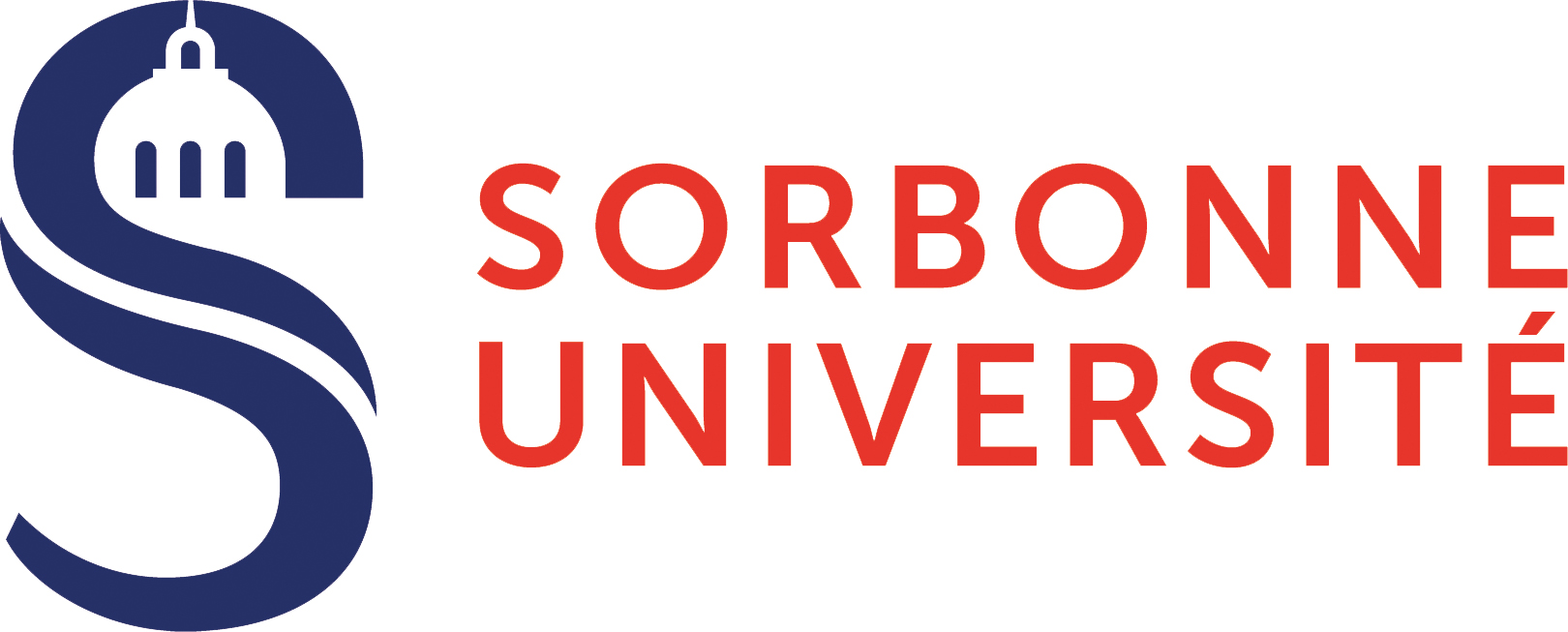}\hspace{1cm}
        \includegraphics[width=0.12\textwidth]{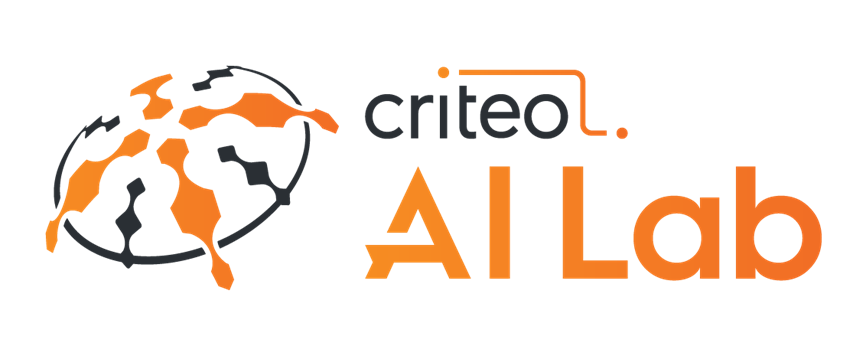}
    \end{figure}
    \vspace{0.5cm}
    \normalsize \textbf{Ismaël Zighed$^{1,2}$, Nicolas Thome$^{2}$, Patrick Gallinari$^{2}$, Taraneh Sayadi$^{3}$} \\ % Example of superscript for affiliations
  { \fontsize{9}{10}\selectfont  
  \textit{$^1$ Institut Jean le Rond d'Alembert - Sorbonne Université, $^2$ ISIR - Sorbonne Université, $^3$ M2N - Conservatoire National des Arts et Métiers} } \\ % Example of superscript for multiple affiliations
  { \fontsize{9}{10}\textit{Corresponding authors: Ismaël Zighed (ismael.zighed@sorbonne-universite.fr), Nicolas Thome (nicolas.thome@sorbonne-universite.fr), Patrick Gallinari (patrick.gallinari@sorbonne-universite.fr), Taraneh Sayadi (taraneh.sayadi@lecnam.net)}}

\end{center}

\begin{abstract}

Reduced order models (ROMs) play a critical role in fluid mechanics by providing low-cost predictions, making them an attractive tool for engineering applications. However, for ROMs to be widely applicable, they must not only generalise well across different regimes, but also provide a measure of confidence in their predictions. While recent data-driven approaches have begun to address nonlinear reduction techniques to improve predictions in transient environments, challenges remain in terms of robustness and parametrisation. 

In this work, we present a nonlinear reduction strategy specifically designed for transient flows that incorporates parametrisation and uncertainty quantification. Our reduction strategy features a variational autoencoder(VAE) that uses variational inference for confidence measurement. We use a latent space transformer that incorporates recent advances in attention mechanisms to predict dynamical systems. Attention’s versatility in learning sequences and capturing their dependence on external parameters enhances generalisation across a wide range of dynamics. Prediction, coupled with confidence, enables more informed decision-making and addresses the need for more robust models. 

In addition, this confidence is used to cost-effectively sample the parameter space, improving model performance \textit{a priori} across the entire parameter space without requiring evaluation data for the entire domain.

\end{abstract}

\end{frontmatter}

\section{Introduction}

The integration of machine learning with physics has fostered a mutually beneficial relationship, driving significant advancements in computational modelling. This synergy has given rise to the development of both physics-informed and physics-inspired models. Physics-informed models, like PINNs~\cite{raissi2017physicsinformeddeeplearning}, explicitly incorporate physical laws into their structure, ensuring that solutions adhere to known scientific principles \cite{cai2021physicsinformedneuralnetworkspinns}. 
On the other hand, physics-inspired models, such as Hamiltonian and Lagrangian networks, are based on the conceptual framework of classical mechanics and other physical theories. These models draw inspiration from the rich mathematical structures found in physics, using concepts like energy conservation, action minimisation, and symmetries to inform their architecture and training procedures. By leveraging these fundamental principles, physics-inspired models are able to encode essential physical behaviours into their design, often leading to enhanced generalisation and more physically plausible predictions. For example, Hamiltonian networks use the principles of Hamiltonian mechanics to model the dynamics of a system, while Lagrangian networks leverage the idea of least action to optimise their predictions ~\cite{Lagrangian, Hamiltonian}. These approaches allow for the modelling of complex physical systems by inherently incorporating dynamical properties and constraints, without requiring explicit equations for every interaction, thus capturing the essence of physical systems in a computationally efficient manner.

Building on the integration of machine learning and physics, the concept of Neural Ordinary Differential Equations (Neural ODEs) has recently emerged as a powerful tool for modelling dynamic systems~\cite{chen2019neuralordinarydifferentialequations}. Neural ODEs provide a continuous-time framework for representing the evolution of dynamic systems, where the time evolution of states is modelled by neural networks in a way that is both smooth and adaptive. This continuous formulation contrasts sharply with traditional discrete-time architectures, such as Residual Networks (ResNets)~\cite{Resnet}, which operate on fixed time steps. Extending these ideas, Neural Operators aim to learn mappings between function spaces rather than finite-dimensional vectors~\cite{DBLP:journals/corr/abs-2108-08481, azizzadenesheli2024neuraloperatorsacceleratingscientific, bahmani2024resolutionindependentneuraloperator}. This innovative approach enables the learning of operators that map between functions, making it highly effective for solving partial differential equations (PDEs) that describe complex systems across different domains. 
 
One notable application of these advanced modelling techniques is Reduced Order Modelling (ROM). ROMs aim to offer computationally efficient alternatives to large, data-intensive models, which is especially valuable for real-time simulations and control in critical systems~\cite{ROM-POD}. By simplifying the system representation, ROMs improve interpretability and transparency, key factors in applications where understanding the underlying mechanics is just as crucial as making accurate predictions. Traditional ROM approaches, such as Proper Orthogonal Decomposition (POD)\cite{POD}, achieve dimensionality reduction by projecting data onto a lower-dimensional manifold based on the overall contribution to the energy norm. An alternative to POD, Dynamic Mode Decomposition (DMD)\cite{DMD}, focuses on decomposing complex systems into the most dominant time-evolving modes. While both POD and DMD, along with their variations~\cite{DMD-1, DMD-2}, have proven effective in identifying coherent structures within data, they are limited by their linear nature, which restricts their ability to fully capture nonlinear dynamics. Consequently, there has been a growing shift towards nonlinear methods~\cite{SSM-1, SSM-2}. In this context, machine learning techniques have been increasingly used to both identify the reduced basis~\cite{Non-LinearProj} and to represent the dynamics within the lower-dimensional manifold~\cite{Luca}. A recent advancement in this area is the machine-learning closure model for POD, referred to as CD-ROM~\cite{MENIER2023115985}, where the projection error orthogonal to the POD basis is addressed by a historical term predicted through a Neural ODE. This methodology offers a more accurate representation of the system’s dynamics, significantly enhancing the performance of reduced order models in capturing the complexities of nonlinear systems.

Koopman theory provides a robust mathematical framework for dealing with complex dynamical systems by assuming the existence of an infinite-dimensional space in which the dynamics are linear. This concept has been applied in various machine learning implementations \cite{doi:10.1137/21M1401243}, but it requires increasing the dimensionality towards infinity, which is often impractical and different from the Reduced Order Model (ROM) framework that our work seeks to integrate. Work combining ROMs with Koopman theory has shown impressive results on complex dynamical problems such as the Navier-Stokes equations \cite{ILED,Gupta2023MoriZwanzigLS}. 

However, these methods have two major limitations. First, they need to be parametrised, or in other words learn and generalise across different operating conditions, referred to in this paper as 'parameters'. Second, they are inherently deterministic, which limits their ability to account for uncertainty in predictions. To address these limitations, we emphasise the importance of adopting a probabilistic framework. Such a framework allows for the quantification of uncertainty, which is critical for providing confidence levels in predictions. A probabilistic approach also offers improved robustness to model inaccuracies and support for risk-aware decision-making. These benefits are critical for fostering confidence and enabling the integration of machine learning models into model-based design.

Recent advancements have focused on integrating probabilistic methods with Reduced Order Models (ROMs) using techniques like Laplace Approximation \cite{Laplace}, Variational Inference \cite{VpROM, vinuesa}, Gaussian Processes \cite{GP}, Ensemble Methods \cite{Ensemble}, and Monte-Carlo Dropout \cite{MCDropout}. However, these methods are seldom applied to non-stationary dynamical systems in practice, largely due to the challenges of integrating them with temporal models such as Neural Ordinary Differential Equations (NODEs), Recurrent Neural Networks (RNNs), or Transformers. An exception is Gaussian Processes, which can model time-series data as multivariate Gaussian distributions, but they come with high computational costs and often require complex tuning of kernel parameters.

To address the difficulty of capturing time dependencies, several approaches have combined traditional reduction methods with Transformers in the latent space \cite{SLT}. Transformers, in particular, have proven highly effective for parametrisation via cross-attention \cite{NIPS2017_3f5ee243}, allowing models to adapt dynamically to different excitation regimes driven by external variables. Additionally, Transformers have shown their versatility and robustness beyond the large language model (LLM) framework, emerging as the state-of-the-art in time-series forecasting. Their generality and flexibility have also facilitated their growing adoption in partial differential equation (PDE) applications, demonstrating their potential to manage and model complex dynamical systems.

Through \textit{UP-dROM}, we aim to contribute to these advances by proposing a Transformer-based, dynamic, parametric and variational ROM. This model is designed to solve unsteady PDEs while considering and adapting to external excitation variables through cross-attention mechanisms, thus exploiting breakthroughs in large language models (LLMs) for PDE applications. In addition, it provides a probabilistic counterpart to most deterministic dynamic modelling baselines, enabling uncertainty quantification. Crucially, our approach achieves this in a computationally efficient manner, significantly reducing the cost compared to Gaussian processes or Bayesian models, while maintaining robustness and scalability for complex dynamical systems.

The paper is structured as follows: Section~\ref{sec:reduction} presents the various steps involved in constructing the model, beginning with the reduction procedure and the corresponding dynamical representation within the reduced manifold. This section also includes the model validation using the test case of a laminar, unsteady flow around a bluff body. The process of parametrising the model is then outlined in Section~\ref{sec:parameter}. Section~\ref{sec:uq} describes the procedure for incorporating uncertainty measures into the model, considering both applications in physical space and parameter space. Section~\ref{sec:adaptive} describes the adaptive procedure, which uses the uncertainty measure to efficiently sample the parameter space. Finally, Section~\ref{sec:conclusion} concludes the paper with a discussion on the findings and directions for future work.

% \subsection{Contribution}
% Model learning dynamlics 
% Generalising 
% UQ 
% Fine tuning method with parameter map

%---------------------------------------------------------
\section{Components of UP-dROM}
\label{sec:reduction}
Consider a dynamical system represented by the equation:
\begin{equation}
    \frac{d\phi}{dt} = F(\phi, \xi, t),
\end{equation}
where \( F \) is a nonlinear operator governing the spatial and temporal evolution of the system, $\phi$ represents the state variables and $\xi$ the parameters governing the dynamical behaviour of the system. The objective is to develop a reduced-order model that accurately represents the dynamics of the given partial differential equation. Considering the high dimensionality of the problem and the inherent presence of coherent structures (modes) in most dynamical systems, we employ dimensionality reduction techniques to both preserve the coherence and reduce the system's dimensionality. To effectively analyse and capture these space-time coherent structures, we propose a space-time strategy that compresses spatial information while simultaneously modelling the temporal dependencies within this lower-dimensional manifold.

\subsection{Dimensionality reduction}
In this work, we reduce the spatial component of the system using a Variational Autoencoder (VAE), which facilitates the compression of high-dimensional physical data into a lower-dimensional nonlinear manifold. The choice of a VAE over a traditional autoencoder is motivated by its ability to incorporate uncertainty quantification, which is a critical aspect of our approach, as discussed in Section~\ref{sec:uq}. The VAE is particularly advantageous because it introduces a probabilistic framework, enabling the model to not only capture the underlying structure of the data but also quantify the uncertainty in the learned representations \cite{VpROM}. This probabilistic nature allows for more reliable inference, especially in situations where uncertainty plays a significant role in prediction or system behaviour. Variational Autoencoders offer the advantage of learning continuous, disentangled latent representations. This structured latent space improves the robustness of inference and enhances generalisation beyond the training distribution, especially when compared to traditional Auto-Encoders. The ability to generalise more effectively is a key factor driving the widespread use of VAEs in generative AI frameworks \cite{VAE}.

Moreover, VAEs and Autoencoders (AEs) have been shown to outperform traditional methods in compressing data from complex physical systems~\cite{ILED,VpROM,kneer:hal-03420320,RONAALP,vinuesa}. This is because VAEs and AEs do not impose strict orthogonality constraints, which is a key limitation in traditional linear compression techniques, such as Proper Orthogonal Decomposition (POD), PCA~\cite{POD,PCA}. In contrast, VAEs and AEs offer the flexibility to relax this constraint, making them better suited for capturing intricate nonlinear structures within the data. Furthermore, the absence of a linearity constraint enables VAEs to generalise more effectively to unseen data distributions. This ability to generalise is particularly important when dealing with real-world physical systems, where the underlying dynamics may vary across different regimes or boundary conditions.

Overall, the use of VAEs provides significant advantages in terms of both data compression and generalisation performance, especially in nonlinear, high-dimensional settings. Their probabilistic nature allows uncertainty quantification, making them a more powerful tool compared to traditional methods that rely on linear approximations. This robustness and flexibility make VAEs a promising approach for modelling and analysing complex physical systems, where the capturing of nonlinear dynamics and the ability to handle uncertainty are crucial for accurate predictions and decision-making. The architecture of the model is shown in Figure~\ref{fig:Main-architecture}. In the provided architecture, \(\mu\) and \(\sigma\) represent the mean and standard deviation of the latent variable distribution, respectively. These parameters define the probabilistic mapping from the input space to the latent space. The determination of each parameter is further explained in Section~\ref{sec:uq}.

\subsection{Temporal evolution}
\begin{figure}
    \centering
    \includegraphics[width=1\linewidth]{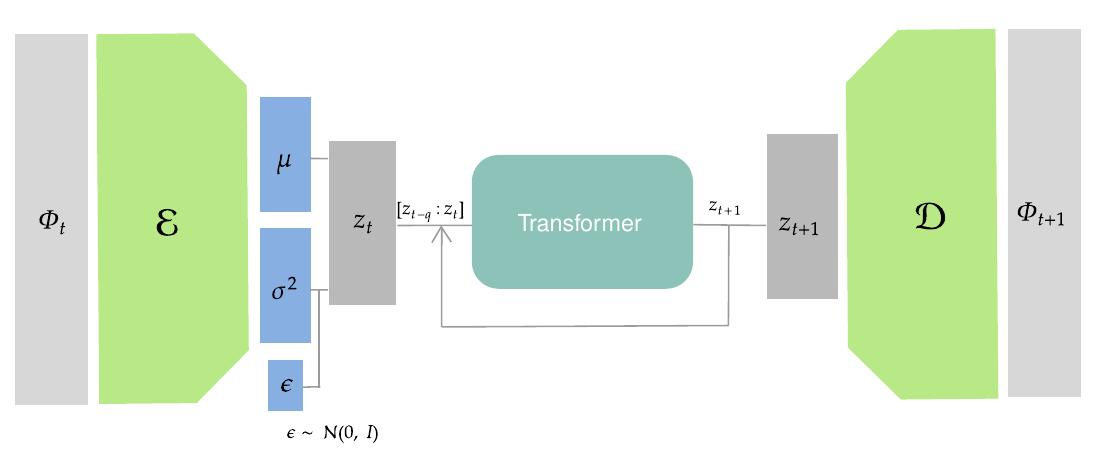}
    \caption{UP-dROM architecture with $q$ the look-back window} 
    \label{fig:Main-architecture}
\end{figure}

To capture the time component of these structures within the latent space, several strategies can be employed. These include utilizing a Koopman operator~\cite{Gupta2023MoriZwanzigLS}, solving the governing equations analytically using Galerkin projection in conjunction with a deep-learning closure model~\cite{MENIER2023115985}, or applying Long Short-Term Memory (LSTM) networks to capture both short-term and long-term temporal dependencies~\cite{LSTM}. In this study, we propose the use of Transformers, similar to those employed in large language models (LLMs)~\cite{NIPS2017_3f5ee243}. This choice is driven by two primary considerations: (i) the attention mechanism in Transformers serves as an effective memory component, facilitating the model’s ability to retain relevant information over time, and (ii) the combination of self-attention and cross-attention mechanisms enables the model to capture complex and nonlinear dependencies, including interactions with external variables. This approach enhances the model’s generalisation capabilities and overall robustness. The resulting architecture is illustrated in Figure~\ref{fig:Main-architecture}.

Regarding point (i), Takens' theorem proves that the state of a dynamical system can be reconstructed from a set of time-delayed embeddings of sparsely sampled observations. The projection performed by the VAE maps the high-dimensional observations \( \Phi \) onto a lower-dimensional latent space \( z \). Time-delayed embeddings constructed from the latent representation \( z \) thus serve as sparse observations of the full state \( \phi \), allowing the application of Takens' theorem directly within the latent space. The trajectory of the system can be uniquely reconstructed from the delayed observations:
\begin{equation}
    z(t), z(t - \tau), z(t - 2\tau), \ldots, z(t - (d-1)\tau)
\end{equation}
By using Takens' theorem, we ensure that the sequence of latent vectors encapsulates the full state of the dynamical system. The transformer, with its attention mechanism, models the interactions and dependencies within this sequence, allowing accurate prediction of future system states and capturing complex temporal dynamics. Similar to how Large Language Models (LLMs) use attention to capture long-range dependencies and contextual relationships in sequential data, transformers in this context exploit time-dependent structures in latent space. This allows the model to generalise effectively, handle non-linearities and adapt to different time scales, thereby increasing robustness and predictive power.

\subsection{Model training}
Transformer and autoencoderare trained together to promote a symbiotic relationship between the two elements, thus fostering an organic modal representation. The total loss is calculated considering a prediction horizon $h$ and a lookback window $q$. The appendix provides further details on the selection of hyperparameters across different test cases, which were chosen through grid search to balance computational efficiency and model accuracy. The loss function is defined as:

\begin{equation}
    \mathcal{L} = \frac{\lambda}{m} (\sum_{k=0}^{m} \| \phi_{n-q:n} -\mathcal{D}\mathcal{E}({\phi}_{n-q:n}) \| + \beta . \text{KLD}) +  \frac{1}{m} \sum_{k=0}^{m} \| z_{n+1:n+h} - \hat{z}_{n+1:n+h} \| + \frac{1}{m} \sum_{k=0}^{m} \| \phi_{n+1:n+h} -\mathcal{D}(\hat{z}_{n+1:n+h}) \|.
\end{equation}
where $\mathcal{E}$ is the encoding process, $\mathcal{D}$ the decoding process, KLD the Kullback-Leibler divergence and $\hat{z}$ the latent variables. In addition, $\lambda$ and $\beta$ are regularisation coefficients. The first term of this loss promotes a mirroring process between encoder and decoder, similar to a classical projection. The Kullback-Leibler divergence term is related to the use of the VAE and is explained in Section~\ref{sec:uq}.

\subsection{Validation of UP-dROM -- application to flow around a cylinder} 
\label{subsec:validation}
The model is validated by predicting the evolution of flow around a blunt object, specifically a cylinder. The flow dynamics are governed by the incompressible Navier-Stokes equations :
\begin{equation}
    \nabla \cdot u = 0
\end{equation}
\begin{equation}
    \frac{\partial u}{\partial t} = - \nabla p - (u \cdot \nabla) u + \frac{1}{Re} \nabla^2 u
\end{equation}
Where \( u \) is the velocity field, \( p \) is the pressure, and \( Re \) is the Reynolds number.
The data are generated using a solver based on the Immersed Boundary Method~\cite{IBMOS}. This solver utilises two grids: a regular Eulerian grid that spans the entire flow domain and a Lagrangian grid that conforms to the boundary of the bluff body, enabling accurate representation of the obstacle-fluid interaction. Once generated, only the Eulerian grid data of the flow field are used; the model is not explicitly informed of the bluff body's position or shape and must learn it implicitly from the data. 

The dataset provided to the model includes all Eulerian grid points and starts from the stationary unstable solution and captures the transient dynamics leading to the stable limit cycle solution. It includes velocities $u$ and $v$ in the $x$ and $y$ directions, respectively, stacked in a state vector $\Phi$. The dataset is divided into a training and a test sets. In this test case, including two crucial phases, to ensure equal representation of the transients and the limit cycle in both the training and test sets, we split the data in half by assigning even time steps to the training set and odd time steps to the test set.

\begin{figure}
\centering
\includegraphics[width=0.75\textwidth]{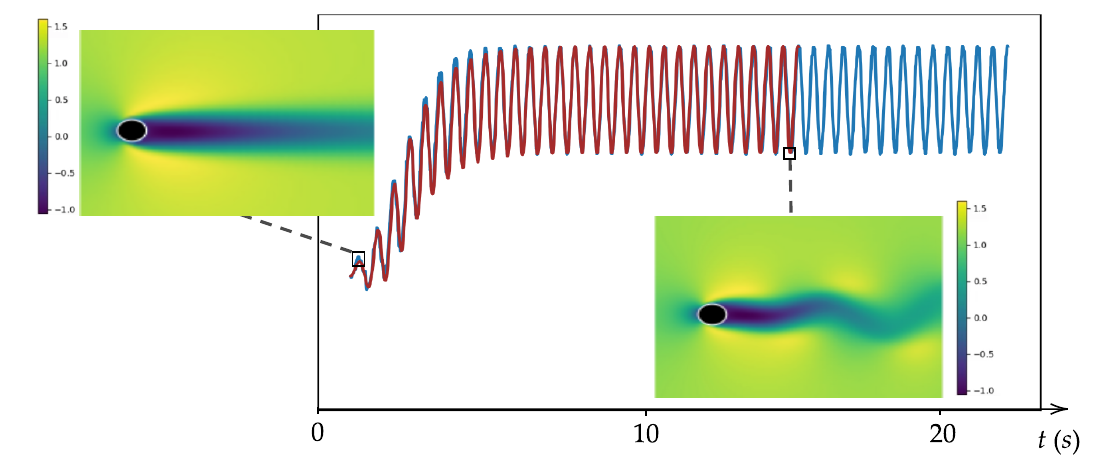}
\caption{Predicted kinetic energy signal using UP-dROM compared with the ground truth.}
\label{fig:Flow_evol}
\centering
\includegraphics[width=0.6\textwidth]{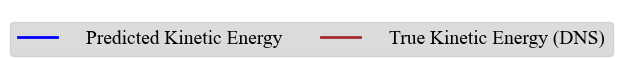}
\end{figure}

The flow behaviour is primarily influenced by the Reynolds number \( R_e \), which is treated here as the external variable or parameter \( \xi \). The Reynolds number is a dimensionless quantity that characterises the ratio of inertial forces to viscous forces in a fluid flow and is defined as:  
\begin{equation}
R_e = \frac{\rho U L}{\mu},
\end{equation}
where \( \rho \) is the fluid density, \( U \) is the characteristic velocity, \( L \) is the characteristic length (such as the length of the bluff body), and \( \mu \) is the dynamic viscosity of the fluid. For this particular bluff body, when \( R_e \) remains below its critical value, the flow closely resembles a stationary solution, with minimal deviations over time. However, beyond this critical Reynolds number, vortex shedding and unsteady flow dynamics emerge in the wake of the cylinder, leading to a more complex and time-dependent behaviour.
Figure~\ref{fig:Flow_evol} shows the base flow provided to the model at the initial time \( t_0 \) and a representative frame at a later time \( t \) when the limit cycle is established. The model should be able to capture the transition dynamics and the stable limit cycle dynamics. 

The transition from the base flow to the stable limit cycle is captured through the evolution of the kinetic energy signal. In Figure~\ref{fig:Flow_evol}, we compare the inferred kinetic energy to the ground truth, followed by a forecasting window that demonstrates the model’s performance. The kinetic energy can be computed from the state vector \( \phi \), which includes both the velocity fields \( u \) and \( v \) over time \( t \), with a spatial discretisation given by \( n_{xy} \). The formula for kinetic energy is given by:

\[
k_t = \frac{1}{2n_{xy}} \sum_{d=1}^{n_{xy}} \left( u_{d,t}^2 + v_{d,t}^2 \right)
\]

This metric effectively averages the spatial components in a way that is physically meaningful. By eliminating the spatial dimensions, it enables us to concentrate on time-dependent signals, making it a valuable visualisation tool for this study.
The results show that the model performs comparably to similar architectures and strategies that use delay embedding to represent the system dynamics in the latent space~\cite{Gupta2023MoriZwanzigLS,MENIER2023115985}.

A key decision in our approach is the preference for a Transformer model over other time-embedding architectures, such as Long Short-Term Memory (LSTM) networks. This choice is informed by recent advancements in embedding memory mechanisms, which significantly enhance the model’s predictive capabilities~\cite{SLT,Chaostheorymeetsdeeplearning, vinuesa, Transformersformodelingphysicalsystems}. Specifically, Transformers leverage the attention mechanism, which allows the model to capture long-range dependencies in the time-series data more effectively than traditional methods. In addition to self-attention, the incorporation of cross-attention mechanisms in Transformers enables the model to respond adaptively to external variables or influences, providing a more comprehensive understanding of the system dynamics. This architecture choice also supports the broader objective of improving model generalisation. By enabling the model to learn relationships and interactions between the data and an external variable such as $R_e$, Transformers are particularly well-suited to address the challenges of generalizing across different conditions, thereby enhancing the robustness and versatility of the model in diverse physical system scenarios.

%---------------------------------------------------------
\section{parametrisation}
\label{sec:parameter}
One of the primary applications of reduced-order models (ROMs) is in multi-query problems, such as control, where a function is evaluated across a wide range of operating conditions to identify the optimal operating point in accordance with a specific design objective. ROMs are particularly advantageous in this context due to their ability to provide cost-effective function evaluations, making them the preferred choice for conducting such analyses. However, for ROMs to be truly effective in these applications, they must maintain predictive accuracy across a range of varying parameters. This requirement represents a significant challenge, as many data-driven reduced-order models applied to dynamical systems fail to satisfy this criterion. 

\begin{figure}[H]
    \centering
        \includegraphics[width=0.75\textwidth]{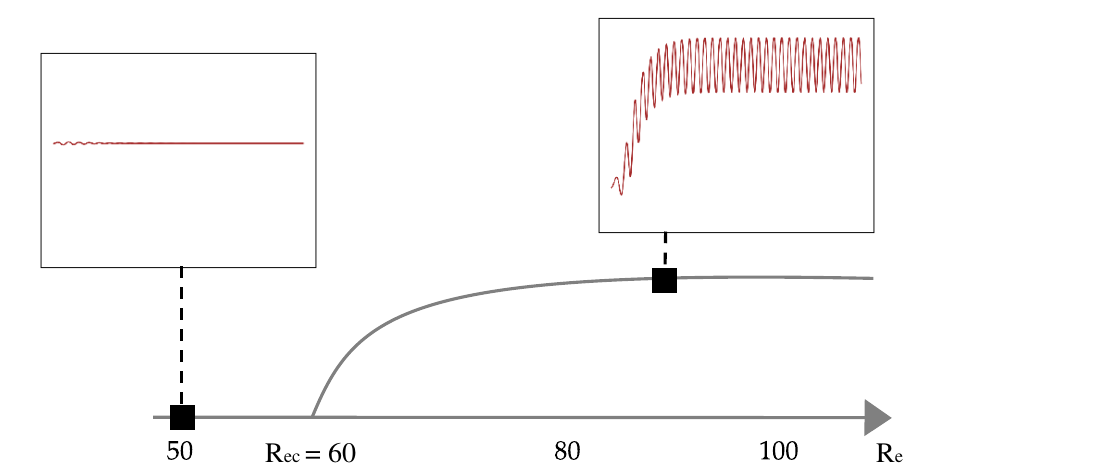}
        \caption{Schematic of a bifurcation diagram with respect to the varying Reynolds number. The two inserted plots are the kinetic energy signals of two simulations performed at $Re=50$ and $Re = 90$, before and after the critical Reynolds number, respectively.}
        \label{fig:bifurcation}
        \centering
        \includegraphics[width=0.45\textwidth]{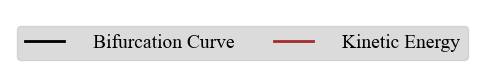}
\end{figure}

Consider the example of flow around a bluff body, as shown in Figure~\ref{fig:Flow_evol}, which serves as a simple dynamical system with a varying Reynolds number, as depicted in Figure~\ref{fig:bifurcation}. As the Reynolds number changes, the system undergoes a bifurcation and transitions into an unstable state. This is evidenced by the steady-state branch bifurcating into oscillatory branches, with unsteady von Kármán vortex street patterns emerging in the flow domain. For the case of 2D flow around a bluff body considered here, we considered an elliptical obstacle with an aspect-ratio $\beta = 1.35$. This aspect-ratio sets the bifurcation at a critical Reynolds Number $R_{ec}\approx60$.

The introduction of a parameter therefore modifies the unstable behaviour of the system, as indicated in the Figure, by eliminating the vortex street, which in turn alters key flow properties such as the drag coefficient. For a reduced-order model to be effective in predicting such a dynamical system, it must be generalizable across different parametric regimes, which often exhibit significantly distinct dynamical behaviours.

\begin{figure}[H]
        \centering
        \includegraphics[width=1\textwidth]{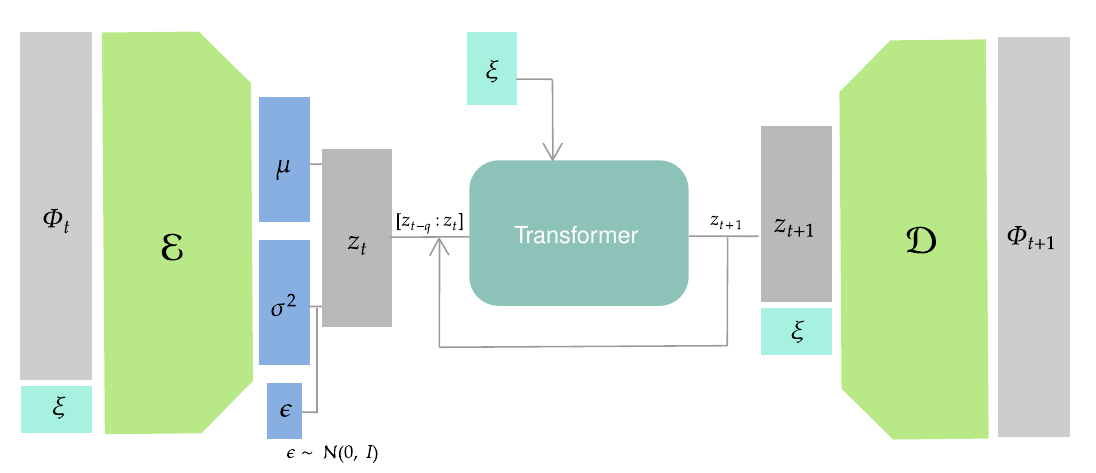}
        \caption{UP-dROM parametrised by external excitation variable $\xi$.}
        \label{fig:p-architecture}
    \end{figure}

The primary objective of this study is to parametrise the model introduced in the previous section and subsequently assess its predictive capabilities across a range of parameter values. To achieve this, we first introduce a parametrisation, denoted as $\xi$, to provide the model with information about external variables that may induce different dynamic response regimes. We incorporate the cross-attention mechanism within the Transformer architecture to account for these external variables during both the learning and inference processes. Specifically, each attention block includes a self-attention component to capture intra-sequence correlations, as well as a cross-attention component to integrate the external parameters. Moreover, to identify an appropriate latent representation for the various dynamic regimes, both the encoder and decoder are informed of these external variables, ensuring robust encoding and decoding across the entire parameter space similar to the approach proposed in ~\cite{VpROM}. The modification to the model architecture is illustrated in Figure~\ref{fig:p-architecture}.

\begin{figure}[H]
        \centering
        \includegraphics[width=0.9\textwidth]{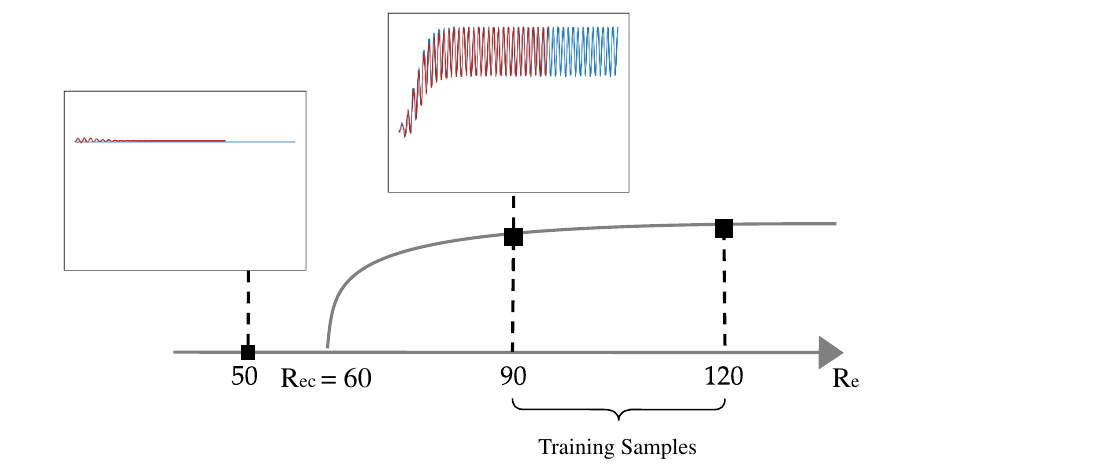}
        \caption{Inferred and true kinetic energy signals in the in-distribution and out-of-distribution training data, superimposed on a sketch of the bifurcation plot with varying Reynolds number. }
        \label{fig:bifurcation_training}
        \centering
        \includegraphics[width=0.7\textwidth]{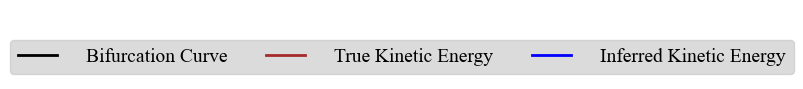}
\end{figure}

To evaluate the model's performance, it was first trained using two Reynolds numbers, $90$ and $120$, both of which correspond to the post-transition limit cycle regime. The model's performance at these two training points is highlighted in Figure~\ref{fig:bifurcation_training}, where the evolution of the total kinetic energy is compared. The Figure demonstrates that the introduction of the parameter does not affect the predictions of the parametrised model, which remain consistent with the performance observed in the previous section. This illustrates the effectiveness of the parametrisation in maintaining accurate inferences.

Additionally, Figure~\ref{fig:bifurcation_training} shows the model's performance when applied to a regime prior to the critical Reynolds number, a regime on which it was not trained. In this regime, the dynamics are significantly different, corresponding to the stable fixed point where the flow is steady and no limit cycle behaviour is observed. This is evident from the evolution of the kinetic energy, which exhibits a decaying and stable trend.

Remarkably, the model captures the underlying dynamics even though it was trained on only a subset of the possible behaviours. For example, it correctly identifies the transition region where the behaviour shifts to a stationary-like state. Therefore, while the overall energy level is not accurately predicted, the model successfully captures the correct dynamical behaviour. In the following section, we propose a strategy to adaptively correct the minor errors in the model's predictions. 

\begin{figure}[H]
    \centering
    \includegraphics[width=0.8\textwidth]{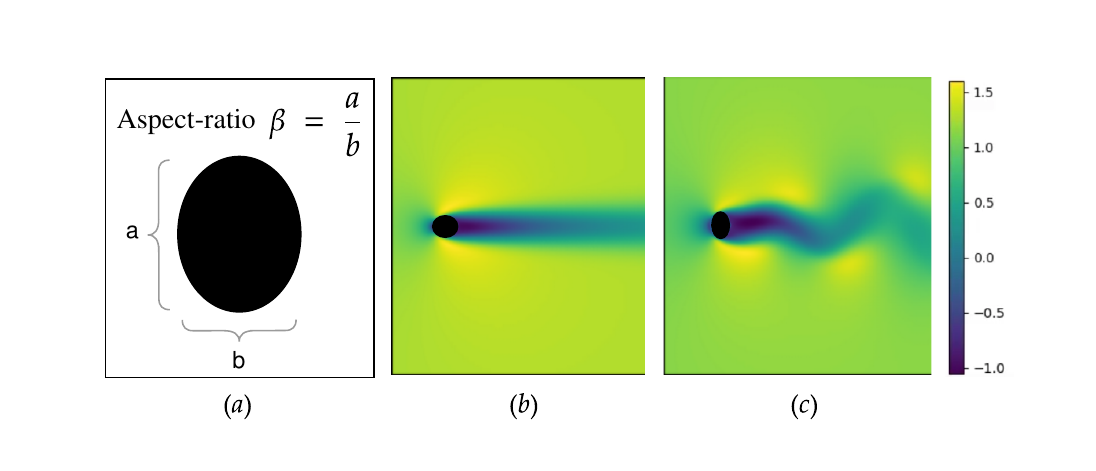}
    \caption{Aspect ratio definition \textit{(a)} and $u$ velocity field at $R_e = 50$ for aspect-ratio  $\beta=1.5$ (\textit{b}), and $\beta=0.75$ (\textit{c}).}
    \label{fig:aspect-ratio}
\end{figure}

To test whether the performance of UP-dROM extends to a larger dimensional parameter space, the aspect ratio of the bluff body was added as a second parameter. The aspect-ratio, as defined in Figure \ref{fig:aspect-ratio}, characterises the shape of the bluff body and significantly influences the flow behaviour. An aspect-ratio smaller than 1 indicates that the obstacle is taller than wide, creating a stronger disruption in the flow. This increased blockage leads to earlier vortex shedding and a bifurcation occurring at a lower Reynolds number.  In contrast, an aspect-ratio greater than 1 corresponds to an elongated body, similar to an airfoil, which disturbs the flow less significantly. In this case, the onset of vortex shedding and bifurcation is delayed, occurring at a higher Reynolds number due to the more streamlined geometry. This effect is illustrated in Figure \ref{fig:aspect-ratio}, where different aspect ratios result in distinct flow regimes, even though the Reynolds number remains unchanged.

\begin{figure}[H]
\centering
\includegraphics[width=1\textwidth]{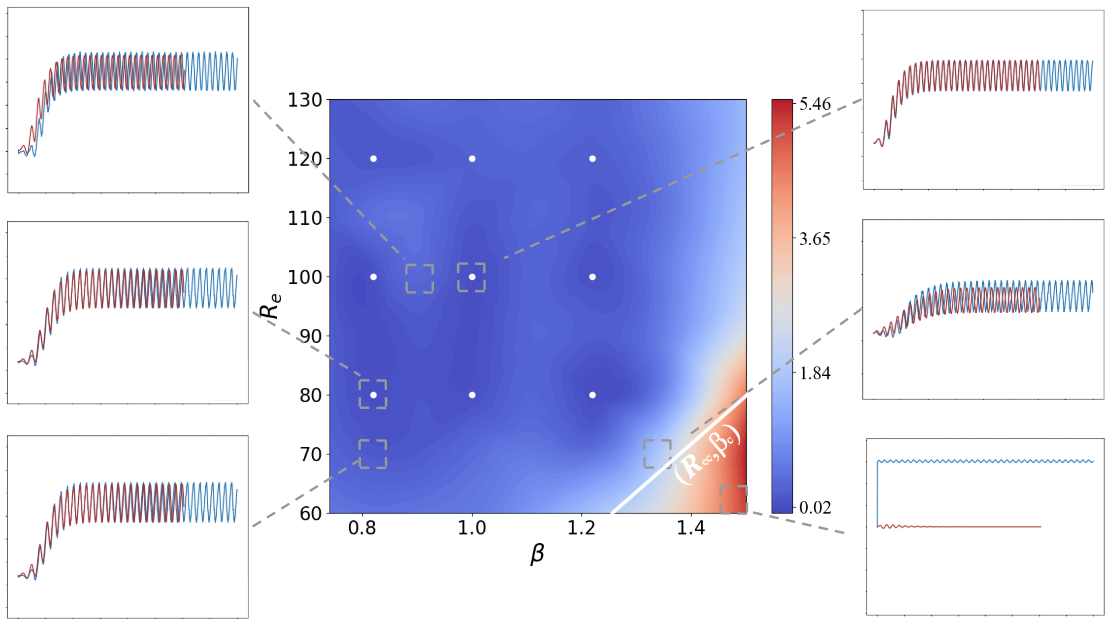}
\caption{ Scaled MSE in the parameter space with predicted \textit{(blue)} vs true \textit{(red)} kinetic energy signals at various location of the $ (R_e, \beta)$ parameter space.}
\vspace{5pt}
    \centering
    \includegraphics[width=1\linewidth]{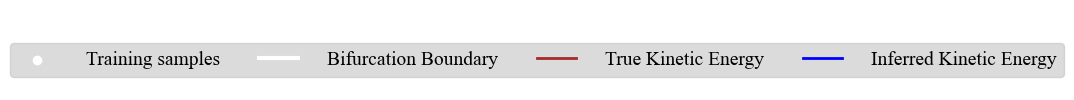}
\label{fig:neutralcurve}
\end{figure}

Performance with a two-variable parametrisation $(R_e, \beta)$ can be visualised using a pointwise scaled MSE, as shown in Figure~\ref{fig:neutralcurve}, which ensures that each grid point is equally weighted in the evaluation. The scaled MSE, computed on test data, is obtained locally using the magnitude of local variations. The resulting heatmap provides a \textit{a-posteriori} perspective on the model's ability to generalise and its overall performance. The signals shown on the heatmap represent the evolution of the kinetic energy for that configuration over time. The model performs well on the training sets, demonstrating its ability to learn and predict. In particular, its interpolation results are strong. However, similar behaviour to the single parameter case is observed. UP-dROM is able to distinguish between stable and limit-cycle dynamics in the two-parameter regime. However, the accuracy of the prediction deteriorates when the model is used in extrapolation mode. 

%---------------------------------------
\section{Uncertainty quantification}
\label{sec:uq}
Most conventional reduction strategies rely on deterministic estimation, where the latent space projection is represented by a single point estimate. This approach fails to capture the inherent uncertainty within the model. In contrast, UP-dROM utilises variational methods to assess uncertainties, providing a probabilistic projection in the latent space instead of a deterministic one. This is achieved through the use of a Variational Autoencoder (VAE). Specifically, given the observed data \( X = D \) (i.e., the dataset \( D \)), the method seeks to infer the posterior distribution \( p(z|X=D) \), where \( z \) represents the latent variables.

The central idea connecting variational inference and uncertainty quantification revolves around approximating the true posterior distribution \( p(z|X=D) \) using a surrogate distribution \( q(z|X=D) \). In this case, the surrogate is chosen to be a Gaussian distribution due to its desirable mathematical properties and ease of interpretation. To ensure that the surrogate closely approximates the true posterior, we minimise the Kullback-Leibler Divergence (KLD) between \( p(z|X=D) \) and \( q(z|X=D) \sim \mathcal{N}(\mu, \sigma^2) \), as:
\[
\text{KLD}(p(z|X=D) \parallel q(z|X=D)) = \int p(z|X=D) \log \frac{p(z|X=D)}{q(z|X=D)} \, dz.
\]

Since \( p(z|X=D) \) is not known \textit{a priori}, we instead utilise the Evidence Lower Bound (ELBO), which can be computed at any stage. Variational inference theory states that minimizing the ELBO, also minimises the KLD between the true posterior and the surrogate distribution \cite{VAE}. The ELBO is given by:
\begin{equation}
    \text{ELBO} = \text{KLD}(p(z) \| q(z|X)) + \mathbb{E}_{z \sim q}[\log(P(X|z))],
\end{equation}
where \( p(z) \) is the prior Gaussian distribution representing the distribution of the latent space. It is typically chosen as \( p(z) \sim \mathcal{N}(0, I) \), which imposes a standard normal prior on the latent variables. This choice ensures that the latent space is continuous and well-structured. It also simplifies the mathematical treatment of the model, making the optimisation more tractable. 
The KLD between two Gaussian distributions such as \( q(z|X=D) \) and \( p(z) \) can be computed as follows,
\begin{equation}
    \text{KLD}(q(z|X=D) \| p(z)) = - \frac{1}{2} \sum_{i=1}^{d} \left( \sigma_i^2 + \mu_i^2 - 1 - \log(\sigma_i^2) \right),
\end{equation}
\[
\text{if} \quad
\left\{
\begin{array}{l}
    q(z|X=D) \sim \mathcal{N}(\mu, \sigma^2) \\
    p(z) \sim \mathcal{N}(0, I)
\end{array}
\right.
\]
The term, \( \mathbb{E}_{z \sim q}[ \log(P(X|z)) ] \), corresponds to the expected log-likelihood of the data, which measures how well the decoder is able to reconstruct the input \( X \) from the latent variable \(z\). It is also often convenient to replace the expectation of the log-likelihood with the Mean Squared Error (MSE) between the predicted and actual values of \( X \), yielding the following approximation:
\[
\mathbb{E}_{z \sim q}[ \log(P(X|z)) ] \approx \text{MSE}(\hat{X}, X),
\]
where \( \hat{X} \) is the reconstructed output of the decoder. 

A Variational Autoencoder (VAE) integrates the Evidence Lower Bound (ELBO) into its loss function, as 
\begin{equation}
    \mathcal{L}_{VAE} = \text{ELBO}.
\end{equation}
 Therefore, the loss function can be computed as,
\begin{equation}
    \mathcal{L}_{VAE} =  \text{MSE}(\hat{X}, X) - \frac{1}{2} \sum_{i=1}^{d} \left( \sigma_i^2 + \mu_i^2 - 1 - \log(\sigma_i^2) \right).
\end{equation}
Through this loss function, the model learns how to better approximate the true posterior distribution using the Gaussian surrogate parametrised by a mean $\mu$ and a variance $\sigma^2$. At the end of the optimisation process, \( q(z|X=D) \) closely approximates the true posterior and provides the density function of the latent projection. Specifically, if the model is confident, \( q(z|X=D) \) will be a narrow Gaussian distribution centred around the expected latent coordinates, indicating low uncertainty in the latent representation. Conversely, if the model is uncertain, \( q(z|X=D) \) will have a wide standard deviation, reflecting higher entropy in the latent space. This broader distribution captures the range of possible latent coordinates consistent with the observed data, effectively quantifying the uncertainty in the projection. \\
During training, the encoder is used at each time step to continuously learn the best-fitting projection of the dynamics at each step. This is not the case during inference, where the dynamics are propagated solely within the latent space using auto-regression. After the initial encoding, the encoder and its associated \( \sigma \) tensor, which captures the uncertainty, do not need to be recomputed, keeping the computational cost low throughout the prediction.

\begin{figure}[H]
\centering
\includegraphics[width=1\textwidth]{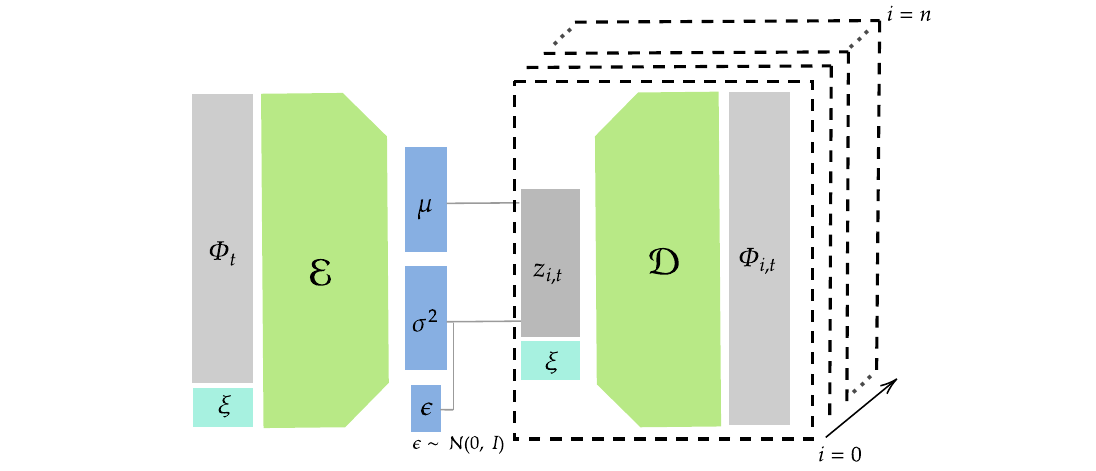}
\caption{UP-dROM Uncertainty Quantification process after inference}
\label{fig:uncertainty}
\end{figure}

We leverage the probabilistic latent projection to compute the uncertainty of the predictions. To do so, a second pass in the model is introduced. In this pass, the predicted dynamic window is processed through the VAE as shown in Figure~\ref{fig:uncertainty}. At each time step, the encoder generates a latent distribution. The method then quantifies the uncertainty by measuring the dispersion $\nu$ of the output sampled $n$ times from this latent distribution. These samples are indexed by $i$ in Figure \ref{fig:uncertainty}. This strategy provides a confidence measure for the entire prediction window, not just the initial state. It also separates the uncertainty estimation from the main forecasting process, allowing for a more comprehensive analysis without increasing the computational burden of the forecasting task. As a result, the strategy remains both robust and scalable, even for large-scale applications. \\
Uncertainty quantification (UQ) using this approach can therefore be interpreted as a stochastic process, where uncertainty is measured by the standard deviation \( \nu \) of the output trajectories over multiple samples from the latent distribution.
Mathematically, the uncertainty $\nu$ at a spatial location \( d \), time \( t \), and parametrisation \( \xi \), given \( n \) samples, is defined as:
\[
\nu_{d,t,\xi} = \sqrt{\frac{1}{n} \sum_{i=1}^{n} \left(\phi_{i,d,t,\xi} - \bar{\phi}_{d,t,\xi} \right)^2}
\]
where \( \bar{\phi}_{d,t,\xi} \) is the ensemble mean. The measure of uncertainty depends on the spatial dimension $d$, the time $t$ and the parametrisation $\xi$. 

\subsection{Uncertainty in space}
\begin{figure}[H]
\centering
\includegraphics[width=1\textwidth]{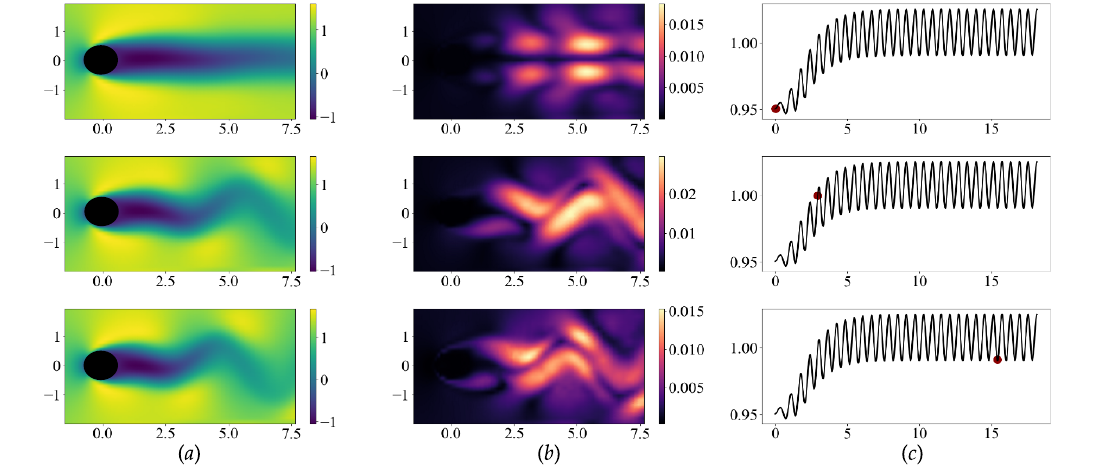}\\[1em]
    \includegraphics[width=1.1\linewidth]{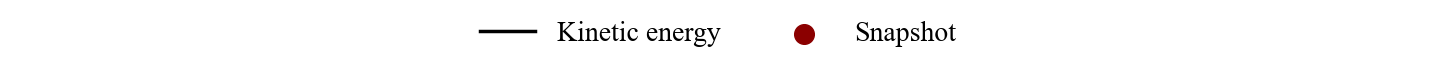}
\caption{U velocity field at the selected snapshots \textit{(a)}, the corresponding uncertainty fields \( \nu_d \) \textit{(b)}, and the snapshot locations indicated on the kinetic energy signal \textit{(c)}, for a fixed parametrisation $\xi$: $R_e = 90$. }
\label{fig:UQ-space-time}
\end{figure}

We can use the uncertainty measure \( \nu_{d,t,\xi} \) to visualise how uncertainty is distributed across the computational domain at different time steps for a fixed parametrisation. Figure~\ref{fig:UQ-space-time}-\textit{a} shows the \( u \)-velocity fields at three key time steps: (1) before the wake forms, during the unstable stationary-like behaviour, (2) during the transition, and (3) once the system has reached the stable limit-cycle, as highlighted in the kinetic energy plot. Figure~\ref{fig:UQ-space-time}-\textit{b} illustrates the distribution of uncertainty in space \( \nu_d \) at these same time steps, computed as:
\[
\nu_{d,t=t_j} = \sqrt{\frac{1}{n} \sum_{i=1}^{n} \left(\phi_{i,d,t=t_j} - \bar{\phi}_{d,t=t_j} \right)^2}.
\]

This figure illustrates a strong correlation between the known space-time dynamics and the uncertainty measurement. Prior to the system's transitioning to the stable limit cycle, uncertainty is evenly distributed along the symmetry axis of the flow domain, as expected, but this distribution breaks down as the dynamics transition. It is in this transient regime that the uncertainty across the entire domain is the highest. The highest confidence is observed upstream of the obstacle, where the dynamics remain relatively stationary throughout the time window. In contrast, lower confidence is concentrated at the edges of the obstacle's wake, suggesting that uncertainty is more closely related to fluctuations in the flow, rather than the instantaneous flow values in specific regions. These fluctuations increase variability within the system, leading to higher uncertainty.

\subsection{Uncertainty in Time}
\begin{figure}[H]
    \centering
    \includegraphics[width=1.0\textwidth]{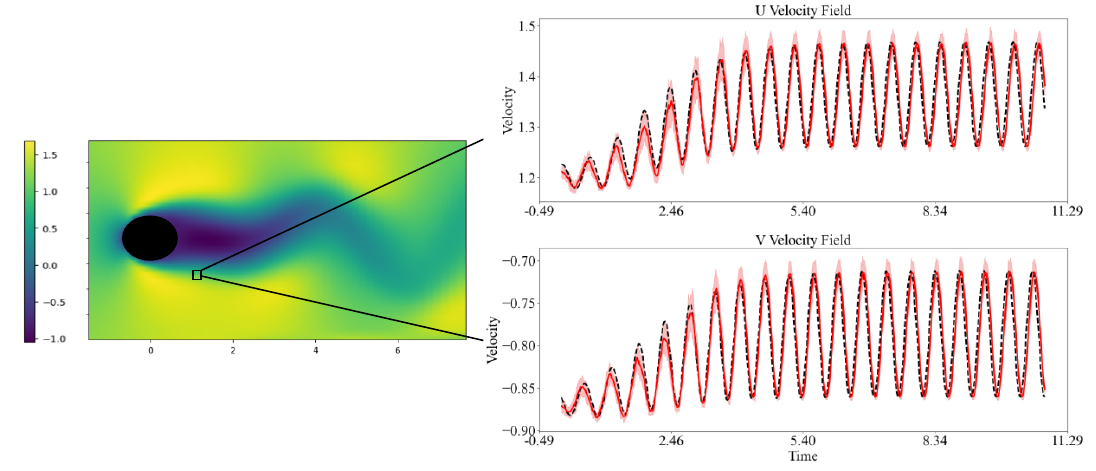}
    \includegraphics[width=1\linewidth]{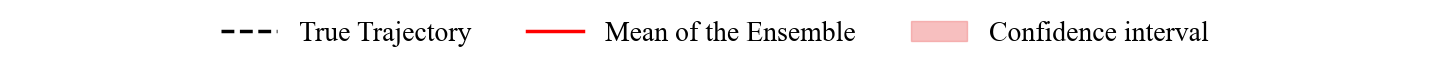}
    \caption{Confidence intervals for velocity fields, at a particular point \( d \) and a specific parametrisation \( \xi \) : $R_e = 90$.}
    \label{fig:prediction-intervals}
\end{figure}

The uncertainty measurement \( \nu_{t,d,\xi} \) can be used to design confidence intervals. By fixing the location at a particular point  and a specific parametrisation, Figure~\ref{fig:prediction-intervals} illustrates such confidence intervals on both velocity fields.

Notably, the true signal lies almost entirely within the confidence interval. It is important to emphasise that this Uncertainty Quantification (UQ) strategy is not time-dependent and is solely driven by the input state. If it were time-dependant, the uncertainty would gradually increase as the dynamic process evolves autoregressively. Instead, we propose a likelihood measure for obtaining a given value, independent of previous values. This approach provides an instantaneous uncertainty measure with minimal computational cost.

As shown in Figure~\ref{fig:prediction-intervals}, the greatest uncertainties occur at the peaks of the signal, corresponding to instances when the local dynamics are most critical. This typically happens when the obstacle's wake shifts within the flow domain, causing substantial changes that the model must predict with accuracy. These transitions trigger increased uncertainty. This behaviour has a significant physical interpretation: uncertainty is highest in regions and time periods where the flow dynamics are most sensitive to changes, such as during sharp transitions, shifts in wake structure, small-scale patterns, and rare events.

\begin{figure}[H]
    \centering
    \includegraphics[width=0.7\textwidth]{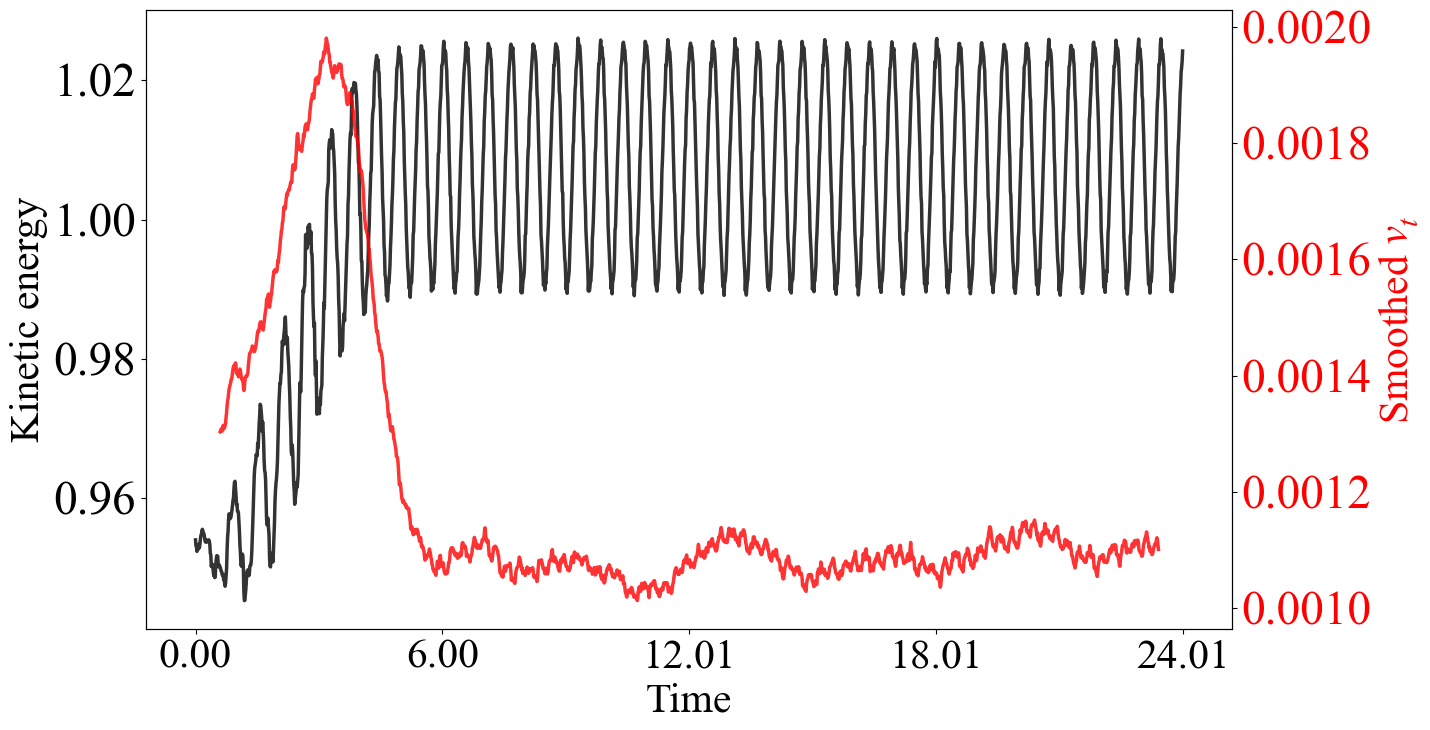}
    \includegraphics[width=1\linewidth]{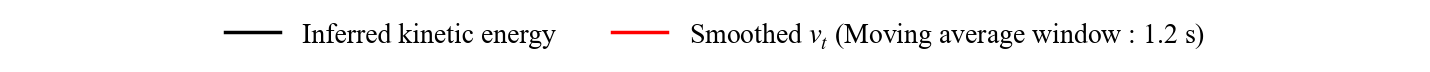}
    \caption{Uncertainty evolution in time.}
    \label{fig:UQ-time}
\end{figure}

Furthermore, as expected, higher uncertainty is observed in the transient part of the signal, compared to the limit cycle. This phenomenon is formally illustrated in Figure~\ref{fig:UQ-time}. To analyse temporal variations in uncertainty for a fixed parameter, we compute the spatially averaged uncertainty \( \nu_{t} \) at each time step \( t \) over \( n_{xy} \) spatial locations,
\begin{equation}
    \nu_{t} = \frac{1}{n_{xy}} \sum_{d=1}^{n_{xy}} \nu_{d,t}
\end{equation}
where,
\begin{equation}
    \nu_{d,t} = \sqrt{\frac{1}{n} \sum_{i=1}^{n} \left(\phi_{i,d,t} - \bar{\phi}_{d,t} \right)^2}.
\end{equation}
The width of the bell curve in Figure~\ref{fig:UQ-time} aligns with the transient phase of the dynamic evolution visible on the kinetic energy signal. Starting from the initial condition, the uncertainty increases as the system transitions towards the limit cycle. Once this stable state is reached, the uncertainty decreases and stabilises at a plateau. This can be explained by two factors: (1) the regime transition induces variability, which like any other source of variability, is associated with lower confidence, and (2) the training distribution contains fewer transient states compared to the limit cycle. As a result, the model has been trained more extensively on the stable state than on the transient phases.

Thus, we highlight two key factors driving uncertainty: the system's inherent variability and the distance from the training distribution. While we have limited control over the first factor, we can influence the second, making it particularly valuable for UQ-driven adaptive training, as demonstrated in Section~\ref{sec:adaptive}.

\subsection{Uncertainty in the Parameter space}
Uncertainty Quantification also provides a confidence measure across the parameter space, aggregating the spatial and temporal components,
\begin{equation}
    \nu_{\xi} = \frac{1}{n_{xy}n_t} \sum_{d=1}^{n_{xy}}\sum_{t=1}^{n_t} \nu_{\xi,d,t},
\end{equation}
where,
\begin{equation}
    \nu_{\xi,d,t} = \sqrt{\frac{1}{n} \sum_{i=1}^{n} \left(\phi_{i,\xi,d,t} - \bar{\phi}_{\xi,d,t} \right)^2},
\end{equation}
with \( n_t \) the time discretisation and \( \bar{\phi}_{\xi,d,t} \) the mean of the ensemble of size \( n \).

\begin{figure}[H]
    \centering
        \includegraphics[width=0.75\textwidth]{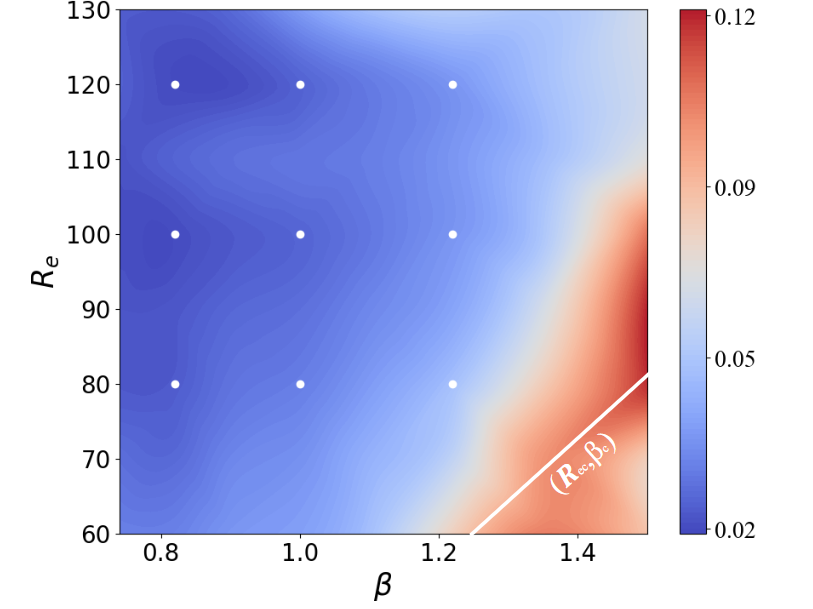} \\[1em]
        \includegraphics[width=0.5\textwidth]{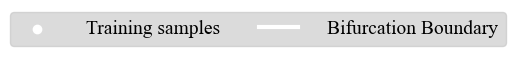}
    \caption{Scaled uncertainty in the \((R_e, \beta)\) parameter space.}
    \label{fig:uq_2D}
\end{figure}

As shown in Figure~\ref{fig:neutralcurve}, the model can learn and infer within the 2D parameter space. Specifically, it can interpolate within the training subspace and extrapolate with varying accuracy depending on the region of the parameter space. The further the inference distribution deviates from the training distribution, the greater the uncertainty of the model, although not uniformly. The direction in which the inference distribution moves away from the training distribution affects the performance of the model. Our uncertainty quantification method can measure this non-Euclidean distance. Figure~\ref{fig:uq_2D} illustrates how our uncertainty quantification process allows us to identify these regions during inference, thus reducing the need for simulation data for evaluation. To improve the performance of the model, it could be retrained using samples from and only from these uncertain regions. An important observation is the strong agreement between the a-posteriori performance evaluation shown in Figure~\ref{fig:neutralcurve} and the a-posteriori inference of these performances. Both highlight the same at-risk dynamic regime. Therefore, our UQ framework is useful for fine-tuning, adaptive training, or error detection.

In particular, the model exhibits lower uncertainty near the training samples, further confirming the correlation between confidence and distance from the training distribution. Furthermore, the model shows particularly low confidence near the bifurcation boundary, demonstrating its strong dynamic awareness. Even without training in the transition regime, the model detects this transition and recognises it as a potential challenge.

We can use this parameter-dependent measure of uncertainty to guide the selection of parameter sets for training. This approach supports adaptive sampling, allowing optimised learning and generalisation for a range of dynamic behaviours, while minimising the amount of training data required. in Section~\ref{sec:adaptive}, we demonstrate a clear correlation between regions of poor model performance and high uncertainty, and identify optimal areas for retraining.

\subsection{Robustness across the parametric regime}
\label{sec:validation_uq}

The performance of the probabilistic model can be assessed using the Relative Mean Squared Error (MSE) calculated using test data, as  
\[
\text{Relative MSE}_\xi = \frac{\sum_{d,t} \left( \hat{\Phi}_{\xi, t, d} - \Phi_{\xi, t, d} \right)^2}{\sum_{d,t} \Phi_{ \xi, d,t}^2} \times 100\%
\]
where, \( \hat{\Phi}_{\xi,d,t}\) is the predicted value for parameter \(\xi\) at time \(t\) and dimension \(d\) and \( \Phi_{ \xi,d,t} \) is the ground truth value for parameter \(\xi\) at time \(t\) and spatial location \(d\).
Another measure is the Continuous Rank Probability Score (CRPS) computed for an ensemble of \(n\) sampled trajectories given by:
\[
\text{CRPS}_\xi =\frac{1}{n}\sum_{i}^{n}\left( \hat{\Phi}_{i,\xi, d,t} - \Phi_{\xi,d,t} \right)^2 - \frac{1}{2n^2} \sum_{i} ^{n}\sum_{j\neq i} ^{n}\left( \hat{\Phi}_{i,\xi,d,t} - \hat{\Phi}_{j,\xi,d,t} \right)^2 .
\]

% The Prediction Interval Coverage Probability (PICP) is the percentage of points in the test set where the true value falls within a prediction interval defined by the mean (\(\mu\)) and three standard deviations (\(\sigma\)) of the ensemble predictions. The interval is chosen as \([ \mu_i - 3\sigma_i, \mu_i + 3\sigma_i ]\), where 3 standard deviations cover approximately 99.7\% of data points assuming a normal distribution.

% For each parameter \(i\), the coverage count is calculated as:

% \[
% \text{PICP}_i =\frac{1}{N} \sum_{t,d} \mathbb{1}_{\left[ \mu_i - 3\sigma_i \leq \Phi_{\text{true}, i, t, d} \leq \mu_i + 3\sigma_i \right]} \times 100\%
% \]

% Where:
% \( \mu_i \) and \( \sigma_i \) are the mean and standard deviation of the ensemble prediction for parameter \(i\),
% \( \Phi_{i, t, d} \) is the true value for parameter \(i\) at time \(t\) and dimension \(d\),
% \( \mathbb{1} \) is the indicator function (1 if true, 0 if false), N is the number of points in the test set.
For a model parametrised by the Reynolds number ($R_e$) and trained for $R_e=90$, $R_e=120$, these two measures are reported in table \ref{tab:result1}. 
\begin{table}[ht]
\centering
\begin{tabular}{|c|c|c|c|c|c|c|c|c|c|c|}
\hline
\textbf{$R_e$ :} & $50$ & $60$ & $70$ & $80$ & \textbf{90} & $100$ & $110$ & \textbf{120} & $130$ & $140$ \\
\hline
Relative MSE (\%) & 0.24 & 0.20 & 1.56 & 1.61 & \textbf{0.28} & 0.71 & 2.49 & \textbf{0.42} & 7.57 & 9.09 \\

\hline
CRPS $(\times 10^{-5})$ & 6.3 & 5.0 & 7.17 & 6.65 & \textbf{5.37} & 3.51 & 6.89 & \textbf{3.22} &  13.7 & 24.5 \\
\hline
% PICP (\%) & 20.01 & 18.78 & 17.23 & 19.30 & \textbf{36.85} & 44.30 & 25.13 & \textbf{49.32} &  10.04 & 6.96 \\
% \hline
\end{tabular}
\caption{Performance metrics evaluated on test data across various Reynolds numbers.}
\label{tab:result1}
\end{table}

The table indicates that with just two training points, the relative error stays below 10\%, and drops even further to below 2.5\% when the Reynolds number \( R_e \) is less than 130, which is within the range of the training distribution. This trend persists even when extrapolating outside the training distribution, particularly for \( R_e < 90 \). As a result, the model maintains satisfactory predictive performance, even in critical applications, highlighting its robustness. Additionally, these findings suggest that the impact of extrapolation on performance is not uniform across the entire parameter space. 

The Continuous Ranked Probability Score (CRPS) is on the order of \( 10^{-4} \), indicating a very low value. Conceptually, CRPS measures the difference between the mean distance of the ensemble from the true target and the variance within the ensemble. A low CRPS indicates that each sampled trajectory is close to the true target trajectory and that the ensemble members themselves are close to each other. Since the model performs well near the training distribution, there is little variation in the ensemble predictions. However, as performance decreases (e.g., when \( R_e > 130 \)), the CRPS increases, reflecting a greater variety within the ensemble. Such a low CRPS is synonymous with an accurate and reliable model.

the performance of a model that has been parametrised by the Reynolds Number ($R_e$) and trained for $R_e=50$, $R_e=60$, $R_e=70$, $R_e=80$, $R_e=90$,$R_e=100$, $R_e=120$, $R_e=140$ is given in table~\ref{tab:result2}, which illustrates the impact of retraining on the model's performance across the parameter space.

\begin{table}[ht]
\centering
\begin{tabular}{|c|c|c|c|c|c|c|c|c|c|c|}
\hline
\textbf{$R_e$ :} & \textbf{50} & \textbf{60} & \textbf{70} & \textbf{80} & \textbf{90} & $100$ & $110$ & \textbf{120} & $130$ & \textbf{140} \\
\hline
Relative MSE ($10^{-2}\%$) & \textbf{0.02} & \textbf{0.13} & \textbf{2.37} & \textbf{3.59} & \textbf{3.58} & 5.99 & 2.74 & \textbf{5.40} & 11.88 & \textbf{18.24} \\

\hline
CRPS $(\times 10^{-5})$ & \textbf{0.32} & \textbf{1.09} & \textbf{2.61} & \textbf{5.59} & \textbf{3.32} & 3.04 & 4.87 & \textbf{5.98} & 11.01 & \textbf{13.23} \\
\hline
% PICP (\%) & \textbf{83.86} & \textbf{53.08} & \textbf{40.95} & \textbf{29.43} & \textbf{45.68} & 37.81 & 31.16 & \textbf{20.32} &  14.87 & \textbf{12.42} \\
% \hline
\end{tabular}
\caption{Performance metrics evaluated on test data across various Reynolds numbers, with training data encompassing a broader range of Reynolds numbers.}
\label{tab:result2}
\end{table}
The results show that expanding the training distribution reduces the relative MSE to below 0.2\% on test data for the entire parameter range of interest, which, to our knowledge, is among the lowest relative MSE reported for the similar benchmarks of the ``flow around a cylinder". The CRPS remains low, but has not decreased as drastically as the relative MSE, indicating that while the model has become more accurate, the diversity of the ensemble has increased. As the dynamic range in the training distribution expands, the model becomes aware of a wider range of possible dynamics, leading to a decrease in overall confidence. However, in Section~\ref{sec:adaptive} we show that the aggregated uncertainty is not as relevant as its distribution over the parameter space.

\section{Adaptive Sampling for Optimised Generalisation}
\label{sec:adaptive}

A parametrised model naturally prompts the question of how to select the appropriate parameters to ensure effective learning and inference across the entire parameter space. Since gathering training data for the full domain is often infeasible, we are typically limited to a small subset of the parameter space. The central issue, therefore, is identifying the minimal set of parameters that will offer the most comprehensive information, enabling the model to generalise effectively across the entire range of parameters.  We aim to use Uncertainty Quantification (UQ) to efficiently explore the parameter space and construct an adaptive sampling strategy. To this end, the high uncertainties identify regions where model performance is suboptimal and indicate the region of the parameter space where data need to be collected. This approach leads to an adaptive sampling technique that adjusts data sampling based on model performance or uncertainty to improve learning efficiency and accuracy. Our approach is implemented during inference of the UQ framework and reduces the reliance on prior knowledge of the parameter space and on the need for validation data throughout the adaptive procedure, which would be necessary with performance/error driven adaptive sampling.
%\cite{adaptive-sampling}.

\begin{figure}[H]
\vspace{-3cm}
    \centering
    \thispagestyle{empty}
    \begin{tabular}{ccc}  

        \begin{subfigure}{0.3\textwidth}
            \includegraphics[width=\linewidth]{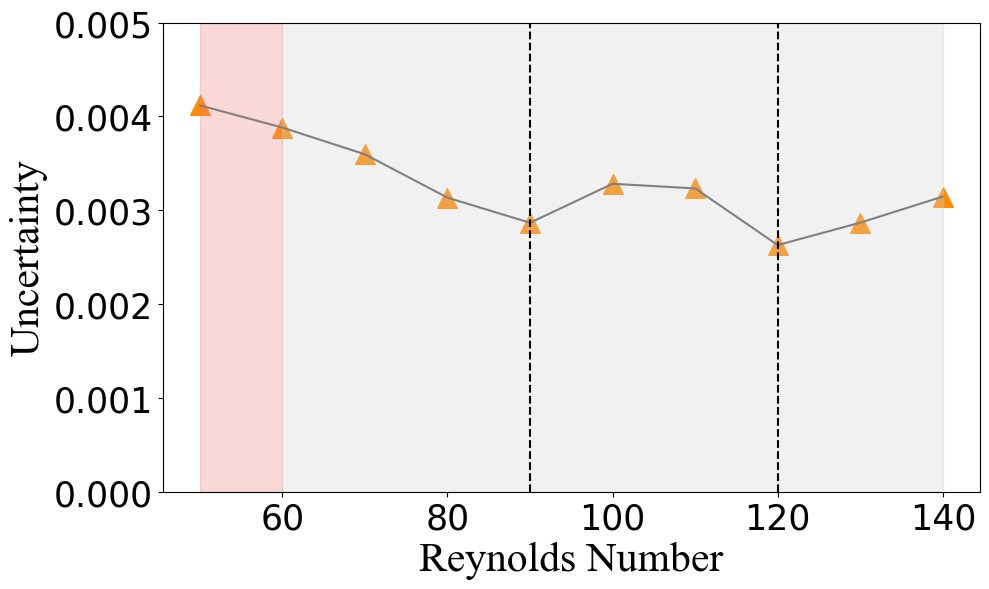}
            \caption{}
            \label{fig:uq_0}
        \end{subfigure} & 
        \begin{subfigure}{0.3\textwidth}
            \includegraphics[width=\linewidth]{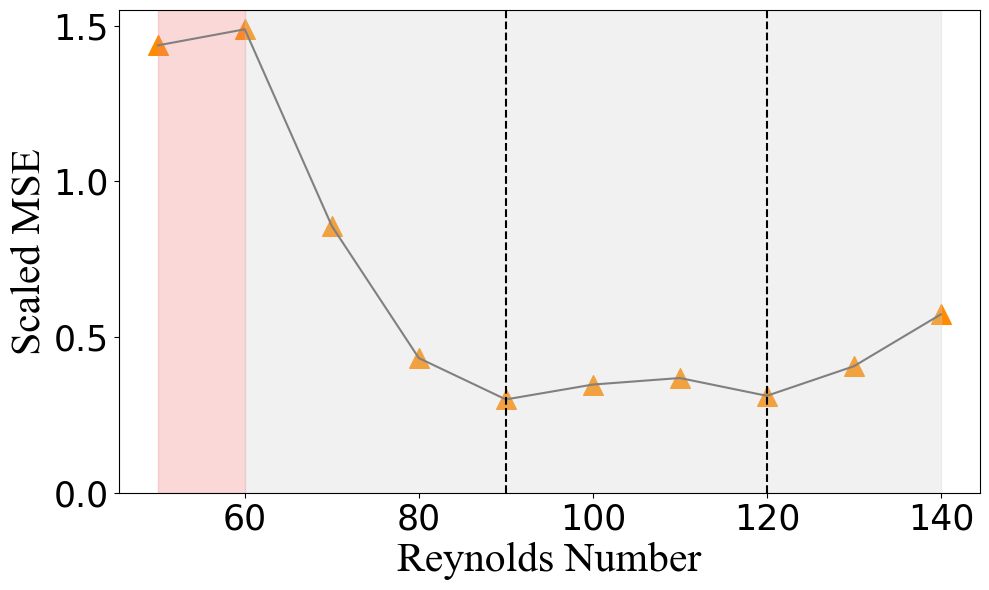}
            \caption{}
            \label{fig:mse_0}
        \end{subfigure} & 
        \begin{subfigure}{0.3\textwidth}
            \includegraphics[width=0.70\linewidth]{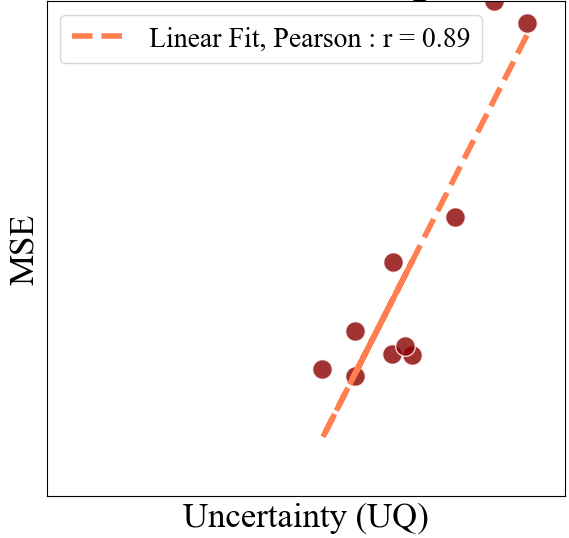}  
            \caption{}
            \label{fig:corr_0}
        \end{subfigure} \\
        \multicolumn{3}{c}{\scriptsize Iteration 0: Initial training} \\

        \begin{subfigure}{0.3\textwidth}
            \includegraphics[width=\linewidth]{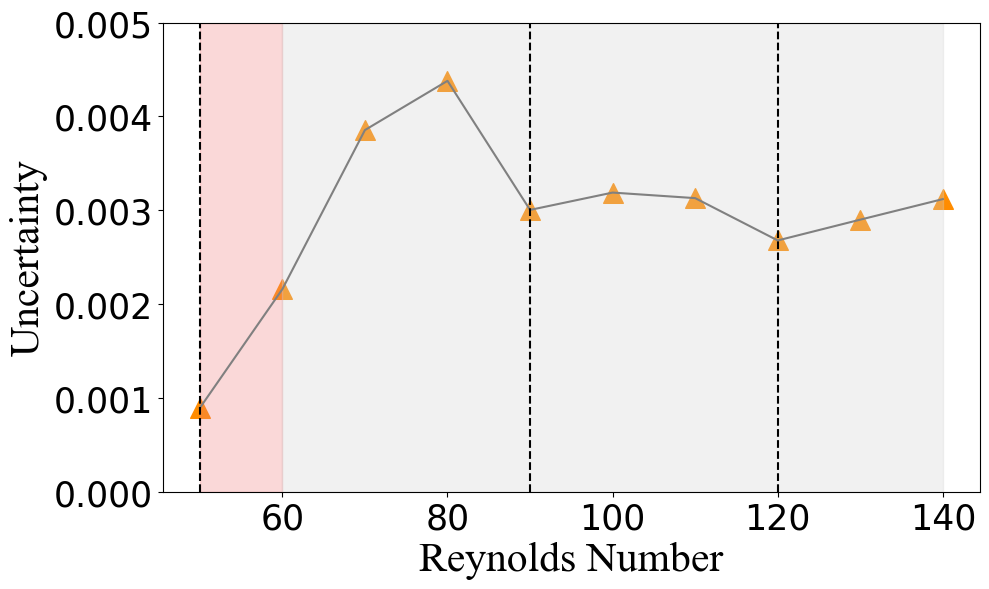}
            \caption{}
            \label{fig:uq_1}
        \end{subfigure} & 
         \begin{subfigure}{0.3\textwidth}
            \includegraphics[width=\linewidth]{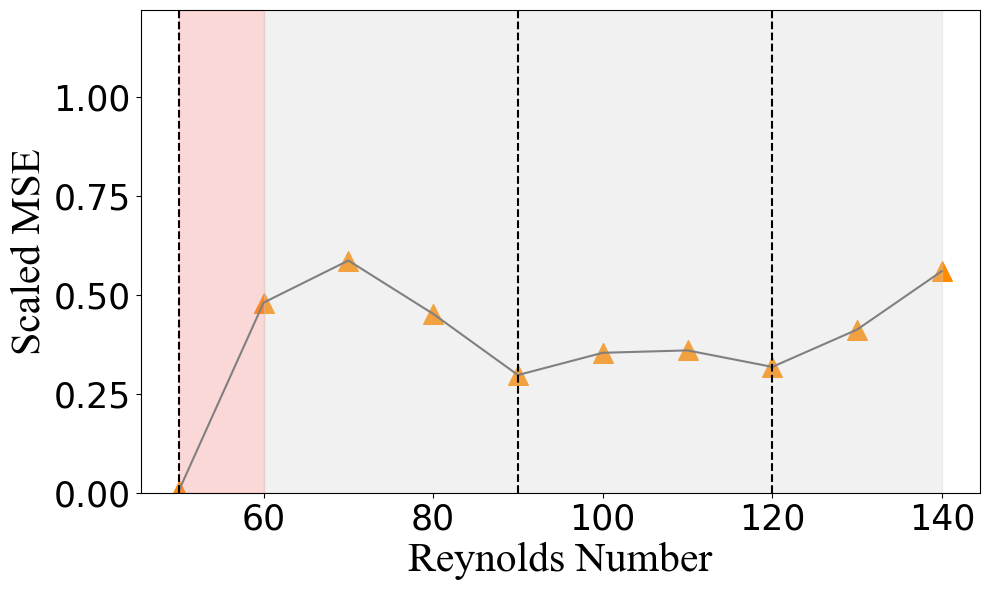}
            \caption{}
            \label{fig:mse_1}
        \end{subfigure} & 
        \begin{subfigure}{0.3\textwidth}
            \includegraphics[width=0.70\linewidth]{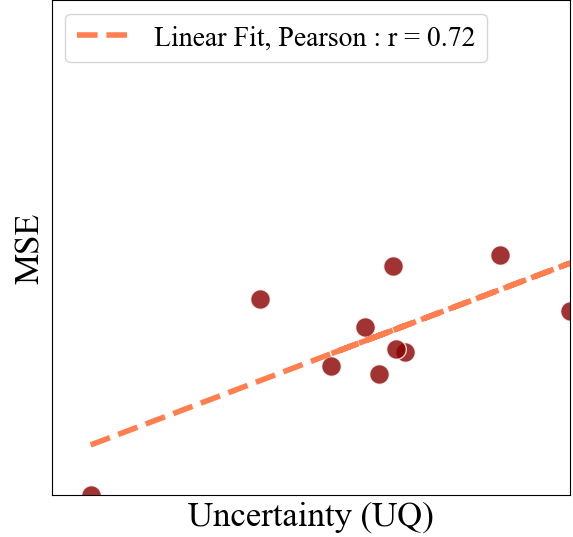}  
            \caption{}
            \label{fig:corr_1}
        \end{subfigure} \\
        \multicolumn{3}{c}{\scriptsize Iteration 1: Retraining at Reynolds number 50} \\

        \begin{subfigure}{0.3\textwidth}
            \includegraphics[width=\linewidth]{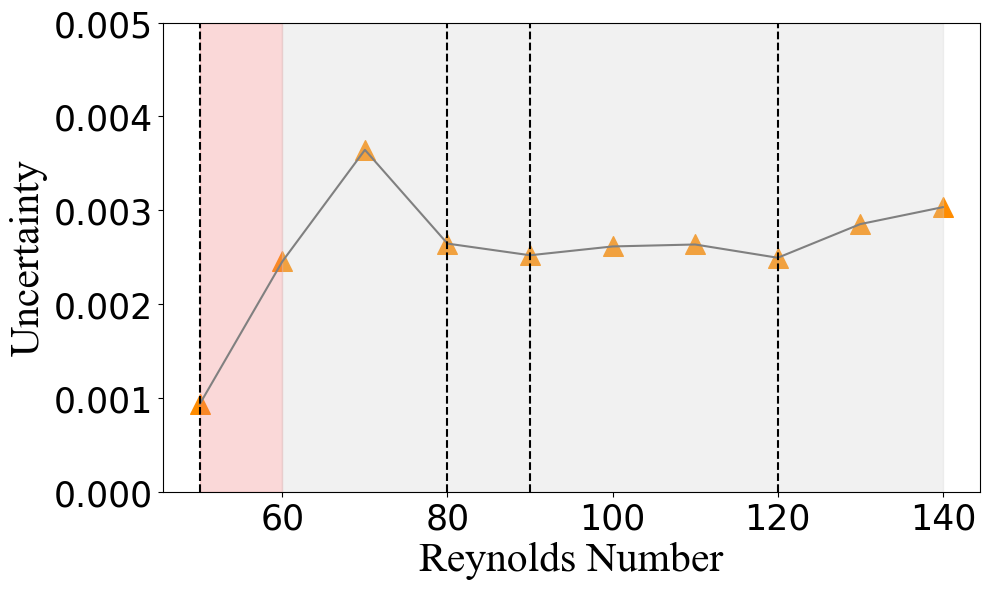}
            \caption{}
            \label{fig:uq_2}
        \end{subfigure} & 
        \begin{subfigure}{0.3\textwidth}
            \includegraphics[width=\linewidth]{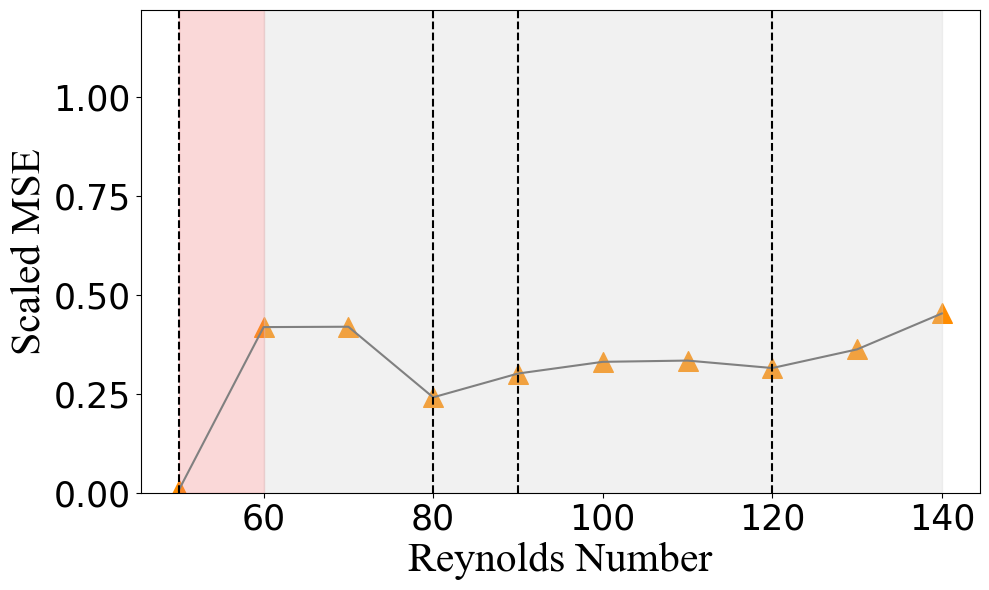}
            \caption{}
            \label{fig:mse_2}
        \end{subfigure} & 
        \begin{subfigure}{0.3\textwidth}
            \includegraphics[width=0.70\linewidth]{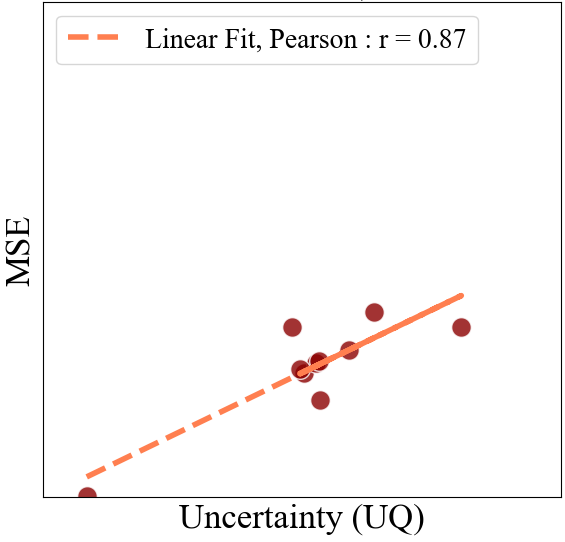}  
            \caption{}
            \label{fig:corr_2}
        \end{subfigure} \\
        \multicolumn{3}{c}{\scriptsize Iteration 2: Retraining at Reynolds number 80} \\

        \begin{subfigure}{0.3\textwidth}
            \includegraphics[width=\linewidth]{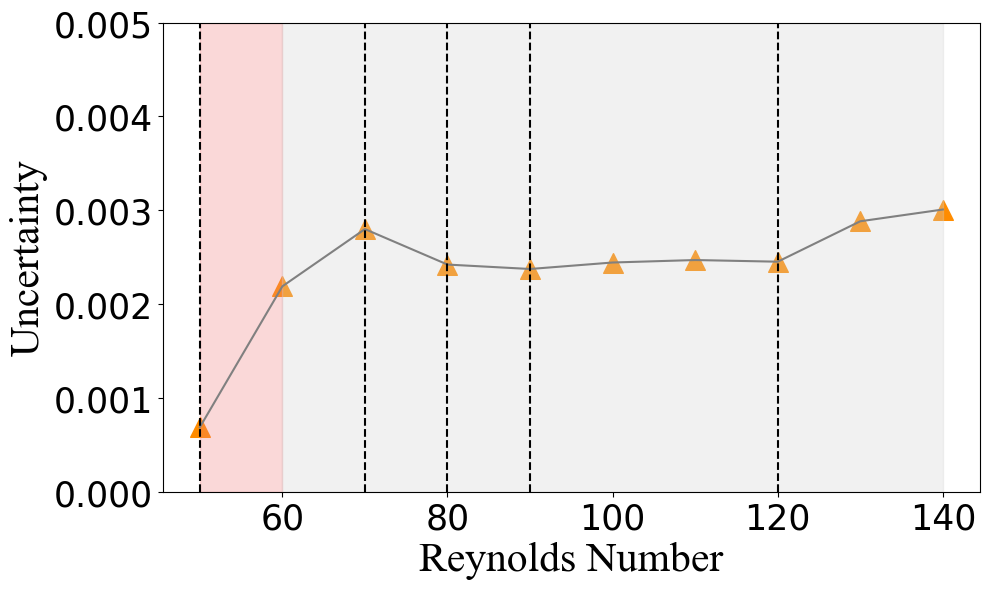}
            \caption{}
            \label{fig:uq_3}
        \end{subfigure} & 
        \begin{subfigure}{0.3\textwidth}
            \includegraphics[width=\linewidth]{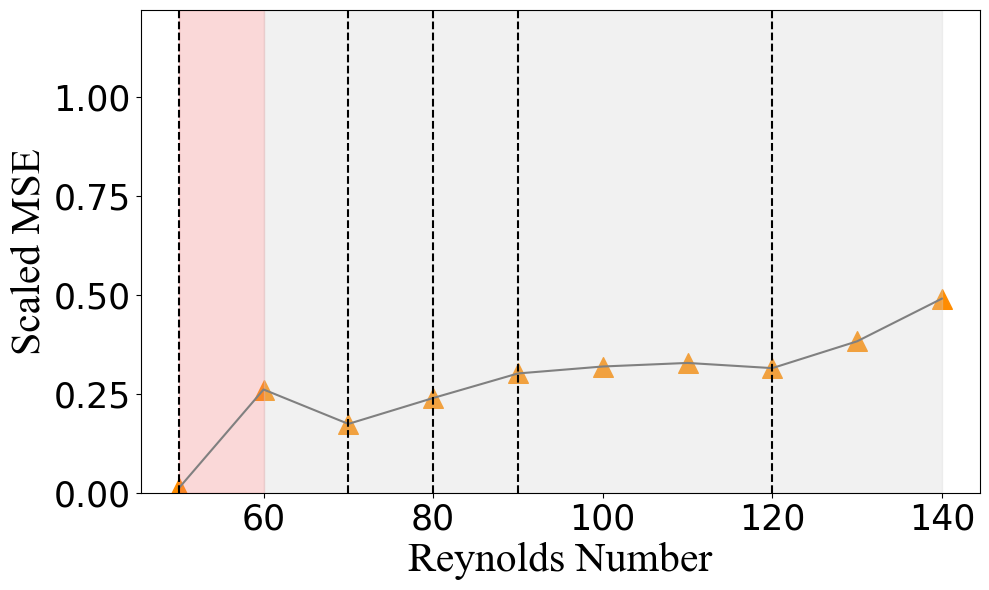}
            \caption{}
            \label{fig:mse_3}
        \end{subfigure} & 
        \begin{subfigure}{0.3\textwidth}
            \includegraphics[width=0.70\linewidth]{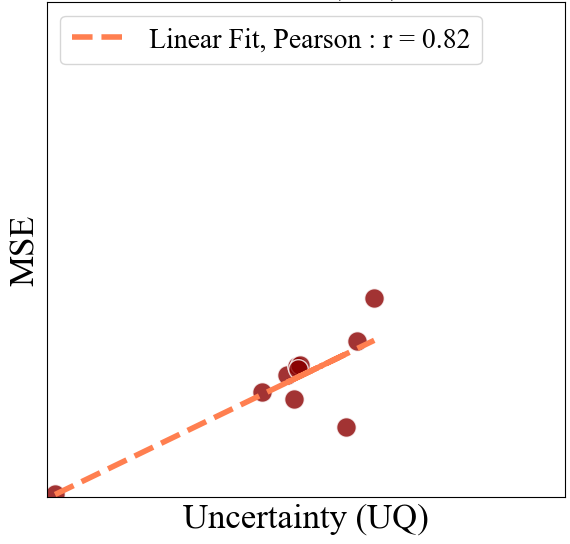}  
            \caption{}
            \label{fig:corr_3}
        \end{subfigure} \\
        \multicolumn{3}{c}{\scriptsize Iteration 3: Retraining at Reynolds number 70} \\

        \begin{subfigure}{0.3\textwidth}
            \includegraphics[width=\linewidth]{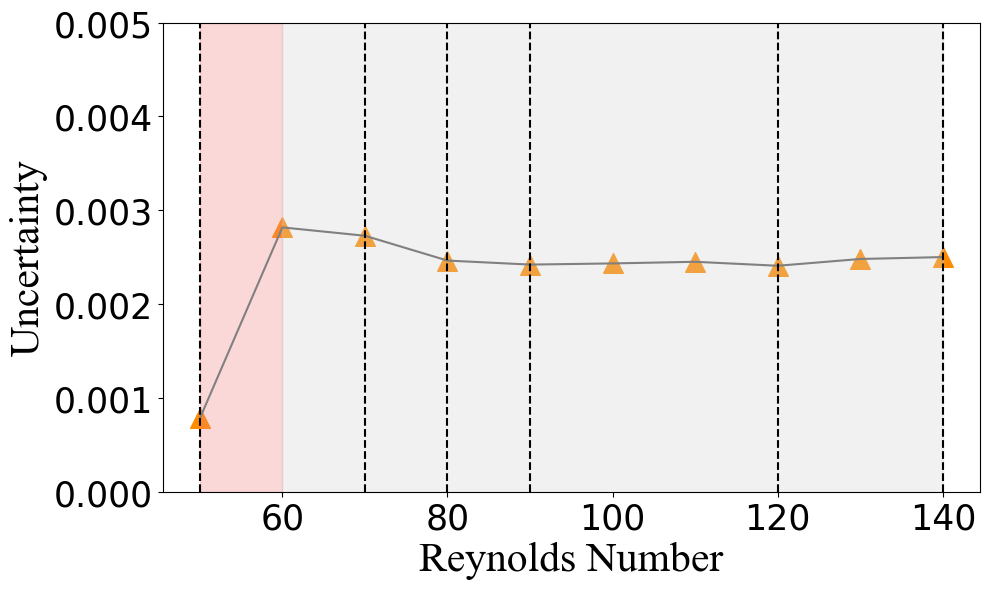}
            \caption{}
            \label{fig:uq_4}
        \end{subfigure} & 
        \begin{subfigure}{0.3\textwidth}
            \includegraphics[width=\linewidth]{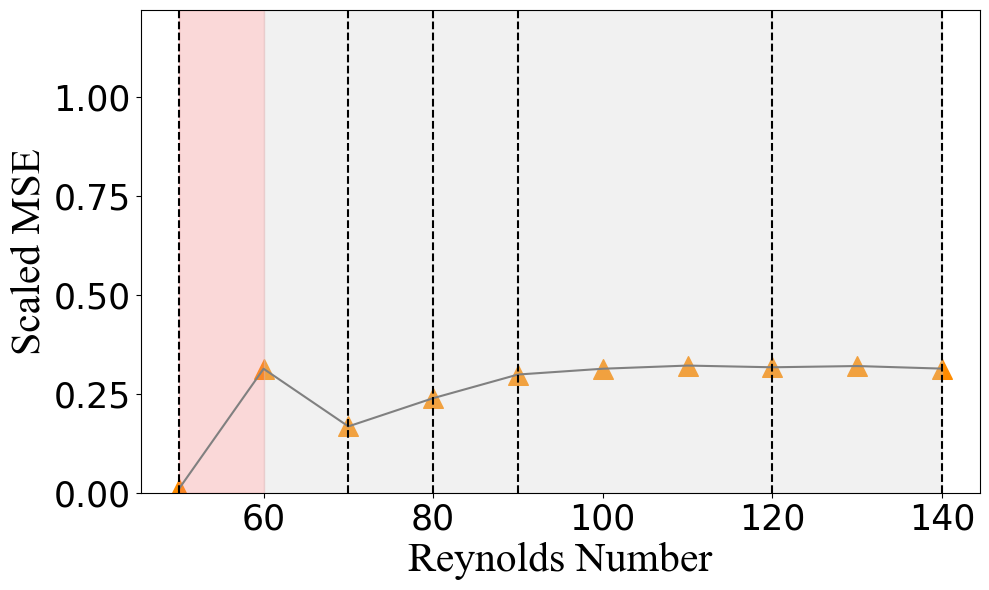}
            \caption{}
            \label{fig:mse_4}
        \end{subfigure} & 
        \begin{subfigure}{0.3\textwidth}
            \includegraphics[width=0.70\linewidth]{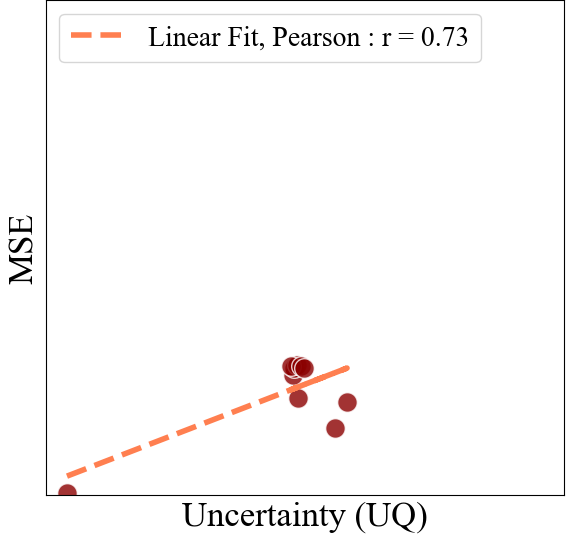}  
            \caption{}
            \label{fig:corr_4}
        \end{subfigure} \\
        \multicolumn{3}{c}{\scriptsize Iteration 4: Retraining at Reynolds number 140} \\

        \begin{subfigure}{0.3\textwidth}
            \includegraphics[width=\linewidth]{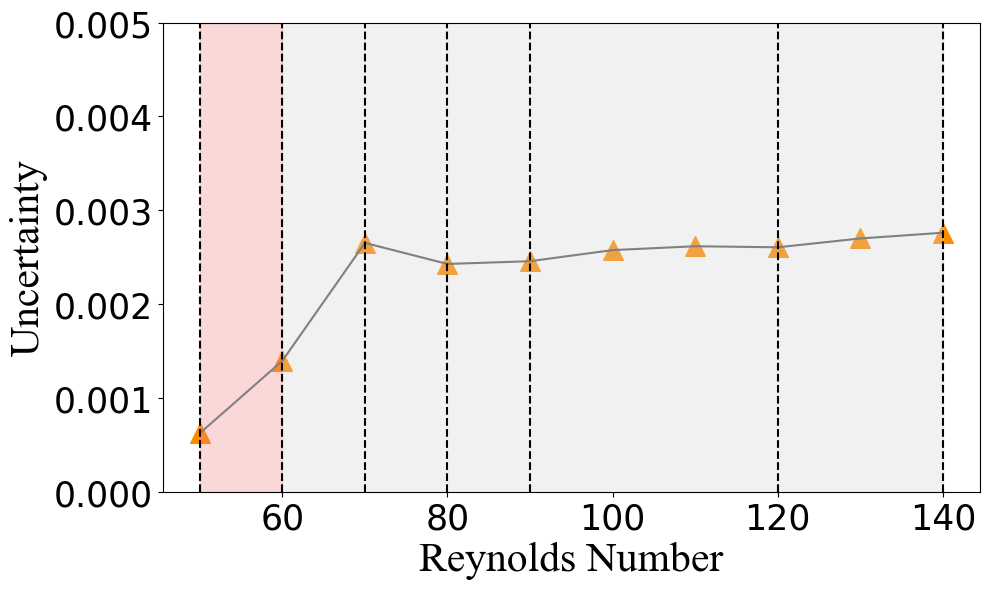}
            \caption{}
            \label{fig:uq_5}
        \end{subfigure} & 
        \begin{subfigure}{0.3\textwidth}
            \includegraphics[width=\linewidth]{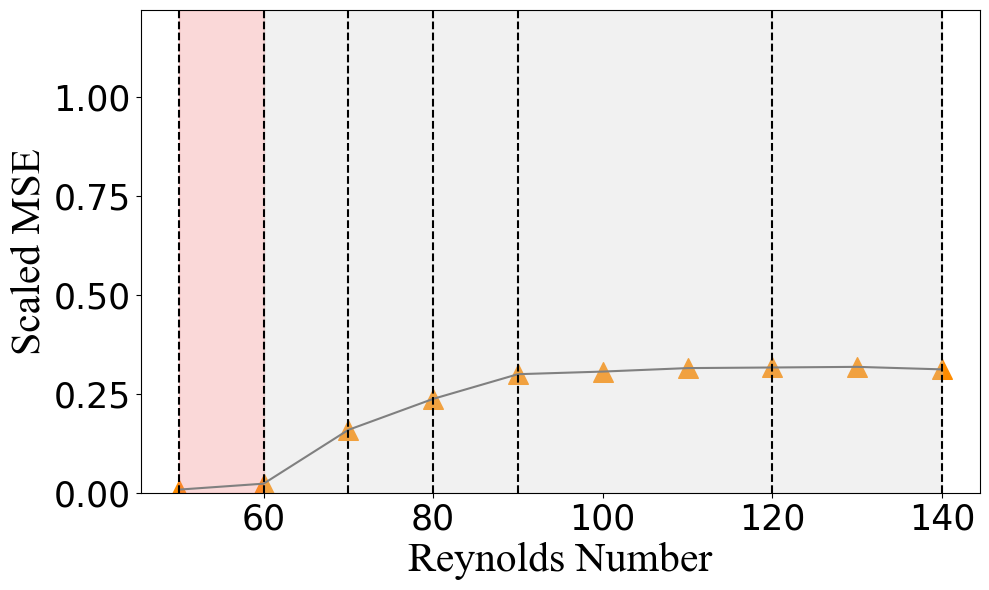}
            \caption{}
            \label{fig:mse_5}
        \end{subfigure} & 
        \begin{subfigure}{0.3\textwidth}
            \includegraphics[width=0.70\linewidth]{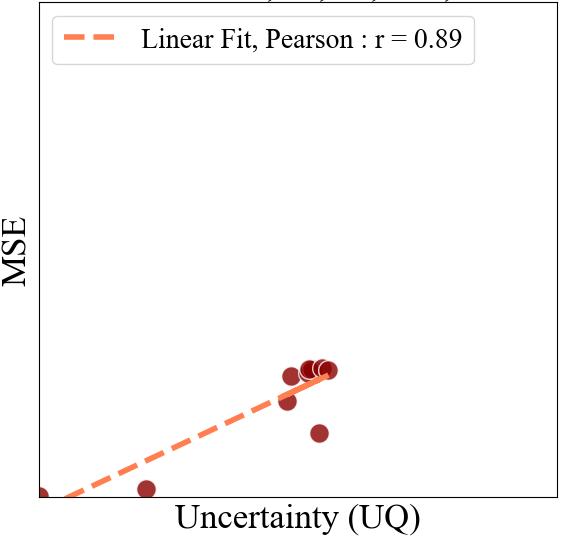} 
            \caption{}
            \label{fig:corr_5}
        \end{subfigure} \\
        \multicolumn{3}{c}{\scriptsize Iteration 5: Retraining at Reynolds number 60} \\

    \end{tabular}
    \caption{Uncertainty, scaled MSE and their correlation throughout a retraining process}
    \label{fig:adaptive-sampling}
    \centering
    \includegraphics[width=\linewidth]{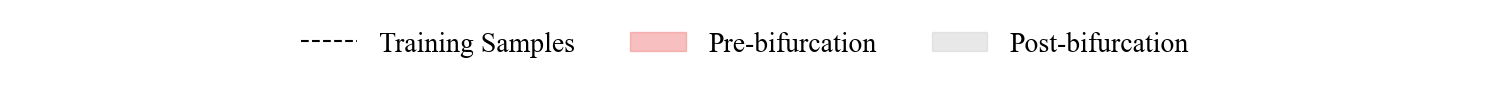}
\end{figure}

This approach can only be effective if the uncertainty can serve as a reliable measure of model performance or predictability. To validate this, we first examine whether there is a strong correlation between the \textit{a posteriori} error and the \textit{a priori} uncertainty. To do this, we choose a parametrised model with a variable Reynolds number $\xi = Re$ and a fixed aspect ratio $(\beta = 1.35)$. This configuration leads to a bifurcation at Reynolds number $\approx 60$. The dynamical regime before this critical Reynolds number $(R_{ec})$ shows a stationary behaviour, while the regime after the bifurcation results in a time-varying limit cycle, as illustrated in Figure \ref{fig:bifurcation} and in Figure \ref{fig:adaptive-sampling} by the different colours of the shaded regions. The experiment is performed over a parameter range from $R_e = 50$ in the pre-bifurcation regime to $R_e = 140$ in the post-bifurcation regime. The model infers over this entire range on 10 equally spaced Reynolds number values, for which validation data are available. We obtain 10 measurements of $\nu_\xi$ that we compare with the model's performance over the same range. Performance is evaluated using the Mean Squared Error (MSE), scaled by the order of magnitude of the signal over the range. Note that in real-world scenarios, validation data over the full range would not be necessary. The scaled MSE is computed here \textit{a posteriori} solely to assess and validate our uncertainty quantification.\\
At initialisation, the model is trained with only two samples in the post-bifurcation regime: $R_e=90$ and $R_e=120$, exposing the model to minimal parameter variety. Figure \ref{fig:adaptive-sampling}-\textit{a} shows the initial uncertainty distribution over the parameter range $\nu_\xi$, while Figure \ref{fig:adaptive-sampling}-\textit{b} shows the scaled error. There is a strong correlation between the two values as formally shown in Figure \ref{fig:adaptive-sampling}-\textit{c}, with the linear fit between $\nu_\xi$ and the error, along with the corresponding Pearson correlation index. In this first step, we observe a Pearson index of $r=0.87 > 0.5$, indicating a strong, positive linear correlation. Not surprisingly, the model is most confident for the in-distribution parameters and specifically for the two training samples. It is least confident in the pre-bifurcation regime. Since the model has never encountered this regime, it exhibits both greater uncertainty and poorer performance. This analysis confirms that the UQ measure serves as a reliable indicator of model predictability or error, and provides the basis for an adaptive sampling strategy that iteratively guides the retraining of the model until uncertainty is minimised over the parameter range of interest.

The uncertainty distribution given by $\nu_\xi$ in Figure \ref{fig:adaptive-sampling}-\textit{a} shows the highest uncertainty at $R_e = 50$. Therefore, we feed the model with this data and perform the first retraining. This retraining is significantly shorter than the initial training, with fewer initial samples included to avoid forgetting. After retraining, both uncertainty and MSE are recomputed at inference and \textit{a posteriori}, respectively, and exhibit a complete redistribution across the parameter range, as shown in Figure \ref{fig:adaptive-sampling}-\textit{d,e}.

After the first retraining, the confidence in the pre-bifurcation segment improves significantly. The stationary-like dynamics make inference easier for the model as the output does not change with time. At this stage, error and uncertainty remain well correlated, as shown in Figure \ref{fig:adaptive-sampling}-\textit{f}. The highest uncertainty now shifts to the other side of the bifurcation, at $R_e = 80$, suggesting the need to refine the model's understanding of the segment near the critical Reynolds number. The data associated with this parameter value becomes the next candidate point for retraining. Interestingly, at this stage, extrapolation beyond the training distribution ($R_e > 120$) is less challenging in terms of both uncertainty and error compared to the lower range ($R_e < 90$).
The retraining process can be extended by iteratively adding samples with the highest uncertainty until the uncertainty converges below a desired value. The complete iterative retraining process, together with the uncertainty distribution $\nu_\xi$,  the scaled MSE and their correlations is shown in Figure~\ref{fig:adaptive-sampling}. The results throughout this process indicate that points with the highest Mean Squared Error (MSE) consistently align with those exhibiting the highest uncertainty. A clear and consistent linear relationship between error and uncertainty is evident. Following each retraining iteration, the Pearson correlation coefficient between the UQ and MSE exceeds 0.7, underscoring a strong positive linear correlation. This demonstrates the effectiveness of our uncertainty quantification method in anticipating poor performance across the parameter space and for adaptive sampling.

\newpage

After completing five retraining iterations, the UP-dROM method demonstrates consistent and uniform learning performance across the entire range of model parameters, effectively capturing the underlying bifurcation dynamics. Notably, the model's uncertainty appears to have converged to a stable final distribution, denoted as $\nu_\xi$. The model progressively identifies parameter points that minimise the error at each training step. As retraining continues at these optimal points, we observe a convergence of both the scaled mean squared error (MSE) and the uncertainty, which both tend towards an optimal distribution.

\section{Conclusions and perspectives}
\label{sec:conclusion}

This paper presents a robust and parametrised reduced-order model (ROM) for time-dependent partial differential equations applied to the Navier-Stokes equations in the laminar regime. The model uses memory embedding and variational inference, first, a nonlinear variational autoencoder is used to reduce the high-dimensional dynamics to a low-dimensional latent space and second, an attention-based transformer captures time dependencies in the reduced manifold. In addition, cross-attention incorporates dependencies on external parameters. This allows UP-dROM to handle different dynamical regimes, making it adaptable to external excitation variables and enhancing generalisation.

Our model demonstrates high accuracy, with a relative mean square error consistently below 0.5\% on the test set for seen parameters. Interpolation between seen parameters remains below 2.5\%, even with limited parameter diversity. Extrapolation is effective in some regions, but less reliable in others. However, the model's Uncertainty Quantification (UQ) framework allows these uncertain regions to be identified \textit{a priori}. A key contribution of this work is the integration of a computationally efficient UQ framework that provides instantaneous confidence estimates  without adding to the computational cost during inference. The approach links uncertainty to both system variability and distance from the training distribution, providing an interpretable measure of confidence across time, spatial domain, and parameter space.

We show that UQ is not only valuable in itself, but also a useful tool for efficiently sampling the parameter space. Since collecting training data across the entire domain is often impractical, training is restricted to a small subset of parameters. Furthermore, we show a strong positive linear correlation between the error and the uncertainty in the parameter space. Therefore, quantifying the uncertainty of the model can guide adaptive sampling, focusing retraining efforts on regions of high uncertainty. This reduces the need for validation data across the entire domain, optimising learning and enhancing generalisation across different dynamical regimes. In addition, the framework effectively captures bifurcation dynamics and complex parametric dependencies, making it well suited for modelling nonlinear fluid systems.

Future work will explore data assimilation strategies to refine the model in uncertain regions. The model can identify regions of high uncertainty in physical space where sparse sensor measurements can enhance the model without requiring additional full-state training data. In addition, using the model's parametrisation as a control variable within a control framework could open new possibilities for flow optimisation.

\newpage

\bibliography{biblio}  

\begin{thebibliography}{10}
\expandafter\ifx\csname url\endcsname\relax
  \def\url#1{\texttt{#1}}\fi
\expandafter\ifx\csname urlprefix\endcsname\relax\def\urlprefix{URL }\fi
\expandafter\ifx\csname href\endcsname\relax
  \def\href#1#2{#2} \def\path#1{#1}\fi

\bibitem{raissi2017physicsinformeddeeplearning}
M.~Raissi, P.~Perdikaris, G.~E. Karniadakis, Physics-informed neural networks: A deep learning framework for solving forward and inverse problems involving nonlinear partial differential equations, Journal of Computational physics 378 (2019) 686--707.

\bibitem{cai2021physicsinformedneuralnetworkspinns}
S.~Cai, Z.~Mao, Z.~Wang, M.~Yin, G.~E. Karniadakis, Physics-informed neural networks (pinns) for fluid mechanics: A review, Acta Mechanica Sinica 37~(12) (2021) 1727--1738.

\bibitem{Lagrangian}
M.~Cranmer, S.~Greydanus, S.~Hoyer, P.~Battaglia, D.~Spergel, S.~Ho, Lagrangian neural networks (03 2020).
\newblock \href {https://doi.org/10.48550/arXiv.2003.04630} {\path{doi:10.48550/arXiv.2003.04630}}.

\bibitem{Hamiltonian}
M.~Mattheakis, D.~Sondak, A.~S. Dogra, P.~Protopapas, \href{https://link.aps.org/doi/10.1103/PhysRevE.105.065305}{Hamiltonian neural networks for solving equations of motion}, Phys. Rev. E 105 (2022) 065305.
\newblock \href {https://doi.org/10.1103/PhysRevE.105.065305} {\path{doi:10.1103/PhysRevE.105.065305}}.
\newline\urlprefix\url{https://link.aps.org/doi/10.1103/PhysRevE.105.065305}

\bibitem{chen2019neuralordinarydifferentialequations}
R.~T. Chen, Y.~Rubanova, J.~Bettencourt, D.~K. Duvenaud, Neural ordinary differential equations, Advances in neural information processing systems 31 (2018).

\bibitem{Resnet}
K.~He, X.~Zhang, S.~Ren, J.~Sun, \href{https://api.semanticscholar.org/CorpusID:206594692}{Deep residual learning for image recognition}, 2016 IEEE Conference on Computer Vision and Pattern Recognition (CVPR) (2015) 770--778.
\newline\urlprefix\url{https://api.semanticscholar.org/CorpusID:206594692}

\bibitem{DBLP:journals/corr/abs-2108-08481}
N.~Kovachki, Z.~Li, B.~Liu, K.~Azizzadenesheli, K.~Bhattacharya, A.~Stuart, A.~Anandkumar, Neural operator: Learning maps between function spaces with applications to pdes, Journal of Machine Learning Research 24~(89) (2023) 1--97.

\bibitem{azizzadenesheli2024neuraloperatorsacceleratingscientific}
K.~Azizzadenesheli, N.~Kovachki, Z.~Li, M.~Liu-Schiaffini, J.~Kossaifi, A.~Anandkumar, Neural operators for accelerating scientific simulations and design, Nature Reviews Physics 6~(5) (2024) 320--328.

\bibitem{bahmani2024resolutionindependentneuraloperator}
B.~Bahmani, S.~Goswami, I.~G. Kevrekidis, M.~D. Shields, A resolution independent neural operator, arXiv preprint arXiv:2407.13010 (2024).

\bibitem{ROM-POD}
V.~Shinde, D.~Gaitonde, Galerkin-pod reduced-order modeling for perturbation analysis and sparse state estimation of compressible flows, 2021.
\newblock \href {https://doi.org/10.2514/6.2021-2603} {\path{doi:10.2514/6.2021-2603}}.

\bibitem{POD}
R.~Pinnau, \href{https://doi.org/10.1007/978-3-540-78841-6_5}{Model Reduction via Proper Orthogonal Decomposition}, Springer Berlin Heidelberg, Berlin, Heidelberg, 2008, pp. 95--109.
\newblock \href {https://doi.org/10.1007/978-3-540-78841-6_5} {\path{doi:10.1007/978-3-540-78841-6_5}}.
\newline\urlprefix\url{https://doi.org/10.1007/978-3-540-78841-6_5}

\bibitem{DMD}
P.~Schmid, J.~Sesterhenn, Dynamic mode decomposition of numerical and experimental data, Journal of Fluid Mechanics 656 (11 2008).
\newblock \href {https://doi.org/10.1017/S0022112010001217} {\path{doi:10.1017/S0022112010001217}}.

\bibitem{DMD-1}
P.~J. SCHMID, Dynamic mode decomposition of numerical and experimental data, Journal of Fluid Mechanics 656 (2010) 5–28.
\newblock \href {https://doi.org/10.1017/S0022112010001217} {\path{doi:10.1017/S0022112010001217}}.

\bibitem{DMD-2}
P.~J. Baddoo, B.~Herrmann, B.~J. McKeon, J.~Nathan~Kutz, S.~L. Brunton, \href{http://doi.org/10.1098/rspa.2022.0576}{Physics-informed dynamic mode decomposition}, Proceedings of the Royal Society A: Mathematical, Physical and Engineering Sciences 479~(2277) (2023) 20220576.
\newblock \href {https://doi.org/10.1098/rspa.2022.0576} {\path{doi:10.1098/rspa.2022.0576}}.
\newline\urlprefix\url{http://doi.org/10.1098/rspa.2022.0576}

\bibitem{SSM-1}
G.~Haller, S.~Ponsioen, \href{https://doi.org/10.1007/s11071-016-2974-z}{Nonlinear normal modes and spectral submanifolds: existence, uniqueness and use in model reduction}, Nonlinear Dynamics 86~(3) (2016) 1493--1534.
\newblock \href {https://doi.org/10.1007/s11071-016-2974-z} {\path{doi:10.1007/s11071-016-2974-z}}.
\newline\urlprefix\url{https://doi.org/10.1007/s11071-016-2974-z}

\bibitem{SSM-2}
G.~Buza, S.~Jain, G.~Haller, \href{http://doi.org/10.1098/rspa.2020.0725}{Using spectral submanifolds for optimal mode selection in nonlinear model reduction}, Proceedings of the Royal Society A: Mathematical, Physical and Engineering Sciences 477 (2021) 20200725.
\newblock \href {https://doi.org/10.1098/rspa.2020.0725} {\path{doi:10.1098/rspa.2020.0725}}.
\newline\urlprefix\url{http://doi.org/10.1098/rspa.2020.0725}

\bibitem{Non-LinearProj}
K.~Lee, K.~T. Carlberg, \href{https://doi.org/10.1016/j.jcp.2019.108973}{Model reduction of dynamical systems on nonlinear manifolds using deep convolutional autoencoders}, J. Comput. Phys. 404~(C) (Mar. 2020).
\newblock \href {https://doi.org/10.1016/j.jcp.2019.108973} {\path{doi:10.1016/j.jcp.2019.108973}}.
\newline\urlprefix\url{https://doi.org/10.1016/j.jcp.2019.108973}

\bibitem{Luca}
Özalp Elise, M.~Luca, \href{https://doi.org/10.1007/s11071-024-10712-w}{Stability analysis of chaotic systems in latent spaces}, Nonlinear Dynamics (2025).
\newblock \href {https://doi.org/10.1007/s11071-024-10712-w} {\path{doi:10.1007/s11071-024-10712-w}}.
\newline\urlprefix\url{https://doi.org/10.1007/s11071-024-10712-w}

\bibitem{MENIER2023115985}
E.~Menier, M.~A. Bucci, M.~Yagoubi, L.~Mathelin, M.~Schoenauer, \href{https://www.sciencedirect.com/science/article/pii/S0045782523001081}{Cd-rom: Complemented deep - reduced order model}, Computer Methods in Applied Mechanics and Engineering 410 (2023) 115985.
\newblock \href {https://doi.org/https://doi.org/10.1016/j.cma.2023.115985} {\path{doi:https://doi.org/10.1016/j.cma.2023.115985}}.
\newline\urlprefix\url{https://www.sciencedirect.com/science/article/pii/S0045782523001081}

\bibitem{doi:10.1137/21M1401243}
S.~L. Brunton, M.~Budi\v{s}i\'{c}, E.~Kaiser, J.~N. Kutz, \href{https://doi.org/10.1137/21M1401243}{Modern koopman theory for dynamical systems}, SIAM Review 64~(2) (2022) 229--340.
\newblock \href {http://arxiv.org/abs/https://doi.org/10.1137/21M1401243} {\path{arXiv:https://doi.org/10.1137/21M1401243}}, \href {https://doi.org/10.1137/21M1401243} {\path{doi:10.1137/21M1401243}}.
\newline\urlprefix\url{https://doi.org/10.1137/21M1401243}

\bibitem{ILED}
E.~Menier, S.~Kaltenbach, M.~Yagoubi, M.~Schoenauer, P.~Koumoutsakos, \href{https://arxiv.org/abs/2309.05812}{Interpretable learning of effective dynamics for multiscale systems} (Submitted).
\newline\urlprefix\url{https://arxiv.org/abs/2309.05812}

\bibitem{Gupta2023MoriZwanzigLS}
P.~Gupta, P.~J. Schmid, D.~Sipp, T.~Sayadi, G.~Rigas, \href{https://api.semanticscholar.org/CorpusID:264172182}{Mori-zwanzig latent space koopman closure for nonlinear autoencoder}, ArXiv abs/2310.10745 (2023).
\newline\urlprefix\url{https://api.semanticscholar.org/CorpusID:264172182}

\bibitem{Laplace}
A.~Kristiadi, M.~Hein, P.~Hennig, \href{https://proceedings.mlr.press/v161/kristiadi21a.html}{Learnable uncertainty under laplace approximations}, in: C.~de~Campos, M.~H. Maathuis (Eds.), Proceedings of the Thirty-Seventh Conference on Uncertainty in Artificial Intelligence, Vol. 161 of Proceedings of Machine Learning Research, PMLR, 2021, pp. 344--353.
\newline\urlprefix\url{https://proceedings.mlr.press/v161/kristiadi21a.html}

\bibitem{VpROM}
T.~Simpson, K.~Vlachas, A.~Garland, N.~Dervilis, E.~Chatzi, Vprom: a novel variational autoencoder-boosted reduced order model for the treatment of parametric dependencies in nonlinear systems, Scientific Reports 14 (03 2024).
\newblock \href {https://doi.org/10.1038/s41598-024-56118-x} {\path{doi:10.1038/s41598-024-56118-x}}.

\bibitem{vinuesa}
A.~Solera-Rico, C.~Sanmiguel~Vila, M.~Gómez-López, Y.~Wang, A.~Almashjary, S.~T.~M. Dawson, R.~Vinuesa, \href{https://doi.org/10.1038/s41467-024-45578-4}{{$\beta$}-variational autoencoders and transformers for reduced-order modelling of fluid flows}, Nature Communications 15~(1) (2024) 1361.
\newblock \href {https://doi.org/10.1038/s41467-024-45578-4} {\path{doi:10.1038/s41467-024-45578-4}}.
\newline\urlprefix\url{https://doi.org/10.1038/s41467-024-45578-4}

\bibitem{GP}
A.~Damianou, Deep gaussian processes and variational propagation of uncertainty, Ph.D. thesis, University of Sheffield (2015).

\bibitem{Ensemble}
T.~G. Dietterich, Ensemble methods in machine learning, in: Multiple Classifier Systems, Springer Berlin Heidelberg, Berlin, Heidelberg, 2000, pp. 1--15.

\bibitem{MCDropout}
Y.~Gal, Z.~Ghahramani, \href{https://proceedings.mlr.press/v48/gal16.html}{Dropout as a bayesian approximation: Representing model uncertainty in deep learning}, in: M.~F. Balcan, K.~Q. Weinberger (Eds.), Proceedings of The 33rd International Conference on Machine Learning, Vol.~48 of Proceedings of Machine Learning Research, PMLR, New York, New York, USA, 2016, pp. 1050--1059.
\newline\urlprefix\url{https://proceedings.mlr.press/v48/gal16.html}

\bibitem{SLT}
I.~Shokar, R.~Kerswell, P.~Haynes, Stochastic latent transformer: Efficient modeling of stochastically forced zonal jets, Journal of Advances in Modeling Earth Systems 16 (06 2024).
\newblock \href {https://doi.org/10.1029/2023MS004177} {\path{doi:10.1029/2023MS004177}}.

\bibitem{NIPS2017_3f5ee243}
A.~Vaswani, N.~Shazeer, N.~Parmar, J.~Uszkoreit, L.~Jones, A.~N. Gomez, L.~u. Kaiser, I.~Polosukhin, \href{https://proceedings.neurips.cc/paper_files/paper/2017/file/3f5ee243547dee91fbd053c1c4a845aa-Paper.pdf}{Attention is all you need}, in: I.~Guyon, U.~V. Luxburg, S.~Bengio, H.~Wallach, R.~Fergus, S.~Vishwanathan, R.~Garnett (Eds.), Advances in Neural Information Processing Systems, Vol.~30, Curran Associates, Inc., 2017.
\newline\urlprefix\url{https://proceedings.neurips.cc/paper_files/paper/2017/file/3f5ee243547dee91fbd053c1c4a845aa-Paper.pdf}

\bibitem{VAE}
L.~Pinheiro~Cinelli, M.~Ara{\'u}jo~Marins, E.~A. Barros~da Silva, S.~Lima~Netto, Variational autoencoder, in: Variational methods for machine learning with applications to deep networks, Springer, 2021, pp. 111--149.

\bibitem{kneer:hal-03420320}
S.~Kneer, T.~Sayadi, D.~Sipp, P.~Schmid, G.~Rigas, Symmetry-aware autoencoders: s-pca and s-nlpca, arXiv preprint arXiv:2111.02893 (2021).

\bibitem{RONAALP}
C.~Scherding, G.~Rigas, D.~Sipp, P.~Schmid, T.~Sayadi, An adaptive learning strategy for surrogate modeling of high-dimensional functions - application to unsteady hypersonic flows in chemical nonequilibrium, Computer Physics Communications 307, publisher Copyright: {\textcopyright} 2024 Elsevier B.V. (Feb. 2025).
\newblock \href {https://doi.org/10.1016/j.cpc.2024.109404} {\path{doi:10.1016/j.cpc.2024.109404}}.

\bibitem{PCA}
B.~Moore, Principal component analysis in linear systems: Controllability, observability, and model reduction, IEEE Transactions on Automatic Control 26~(1) (1981) 17--32.
\newblock \href {https://doi.org/10.1109/TAC.1981.1102568} {\path{doi:10.1109/TAC.1981.1102568}}.

\bibitem{LSTM}
A.~Mavi, A.~C. Bekar, E.~Haghighat, E.~Madenci, \href{https://www.sciencedirect.com/science/article/pii/S0045782523000671}{An unsupervised latent/output physics-informed convolutional-lstm network for solving partial differential equations using peridynamic differential operator}, Computer Methods in Applied Mechanics and Engineering 407 (2023) 115944.
\newblock \href {https://doi.org/https://doi.org/10.1016/j.cma.2023.115944} {\path{doi:https://doi.org/10.1016/j.cma.2023.115944}}.
\newline\urlprefix\url{https://www.sciencedirect.com/science/article/pii/S0045782523000671}

\bibitem{IBMOS}
M.~F. de~Pando, \href{https://doi.org/10.5281/zenodo.3757783}{miguelfp/ibmos: Initial release} (Apr. 2020).
\newblock \href {https://doi.org/10.5281/zenodo.3757783} {\path{doi:10.5281/zenodo.3757783}}.
\newline\urlprefix\url{https://doi.org/10.5281/zenodo.3757783}

\bibitem{Chaostheorymeetsdeeplearning}
B.~Jia, H.~Wu, K.~Guo, \href{https://www.sciencedirect.com/science/article/pii/S0957417424014003}{Chaos theory meets deep learning: A new approach to time series forecasting}, Expert Systems with Applications 255 (2024) 124533.
\newblock \href {https://doi.org/https://doi.org/10.1016/j.eswa.2024.124533} {\path{doi:https://doi.org/10.1016/j.eswa.2024.124533}}.
\newline\urlprefix\url{https://www.sciencedirect.com/science/article/pii/S0957417424014003}

\bibitem{Transformersformodelingphysicalsystems}
N.~Geneva, N.~Zabaras, \href{https://www.sciencedirect.com/science/article/pii/S0893608021004500}{Transformers for modeling physical systems}, Neural Networks 146 (2022) 272--289.
\newblock \href {https://doi.org/https://doi.org/10.1016/j.neunet.2021.11.022} {\path{doi:https://doi.org/10.1016/j.neunet.2021.11.022}}.
\newline\urlprefix\url{https://www.sciencedirect.com/science/article/pii/S0893608021004500}

\end{thebibliography}

\newpage

\section{Appendix}
\subsection{Generalisation to Different Physics}
\begin{figure}[H]
    \centering
    \includegraphics[width=0.5\textwidth]{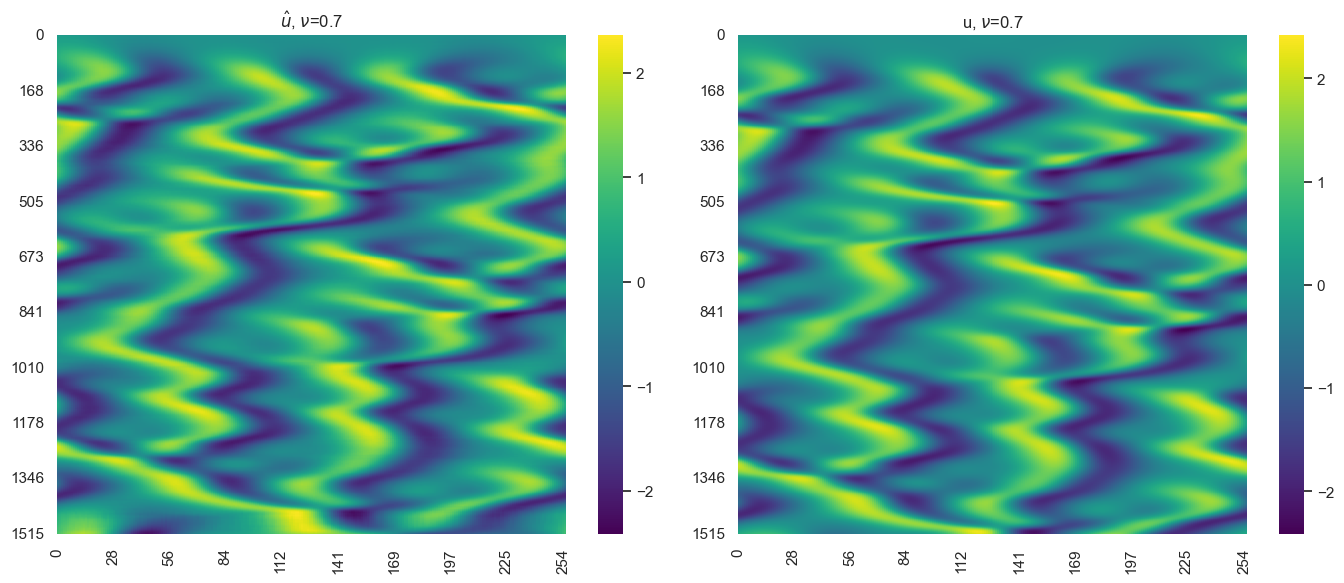} \\
    \includegraphics[width=0.5\textwidth]{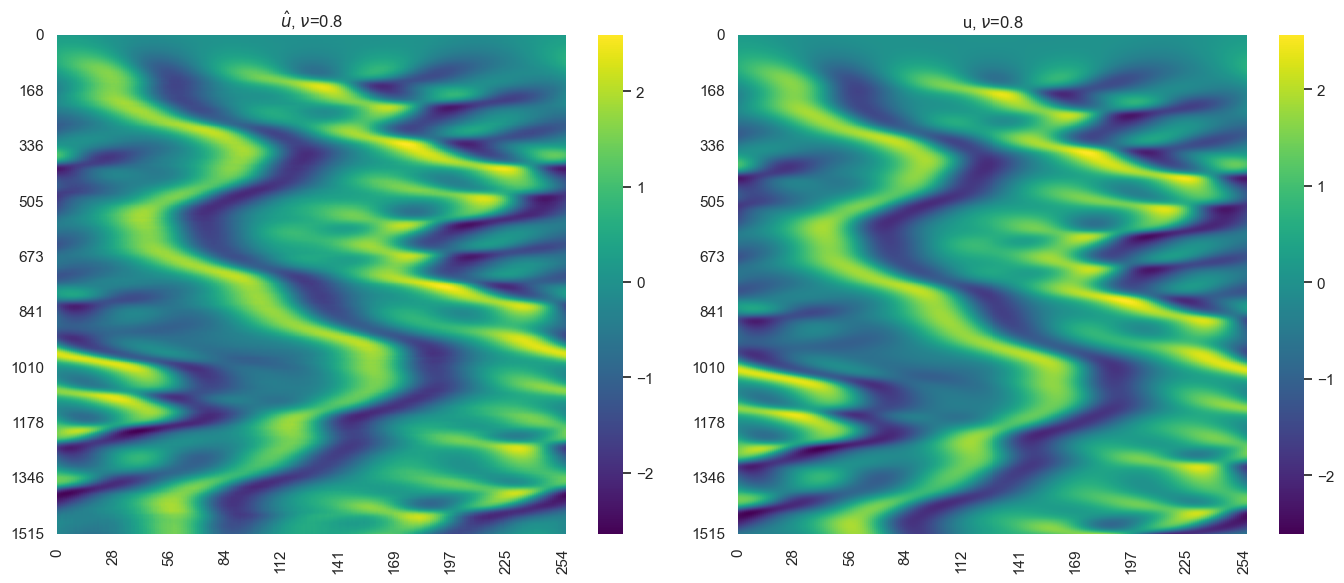} \\
    \includegraphics[width=0.5\textwidth]{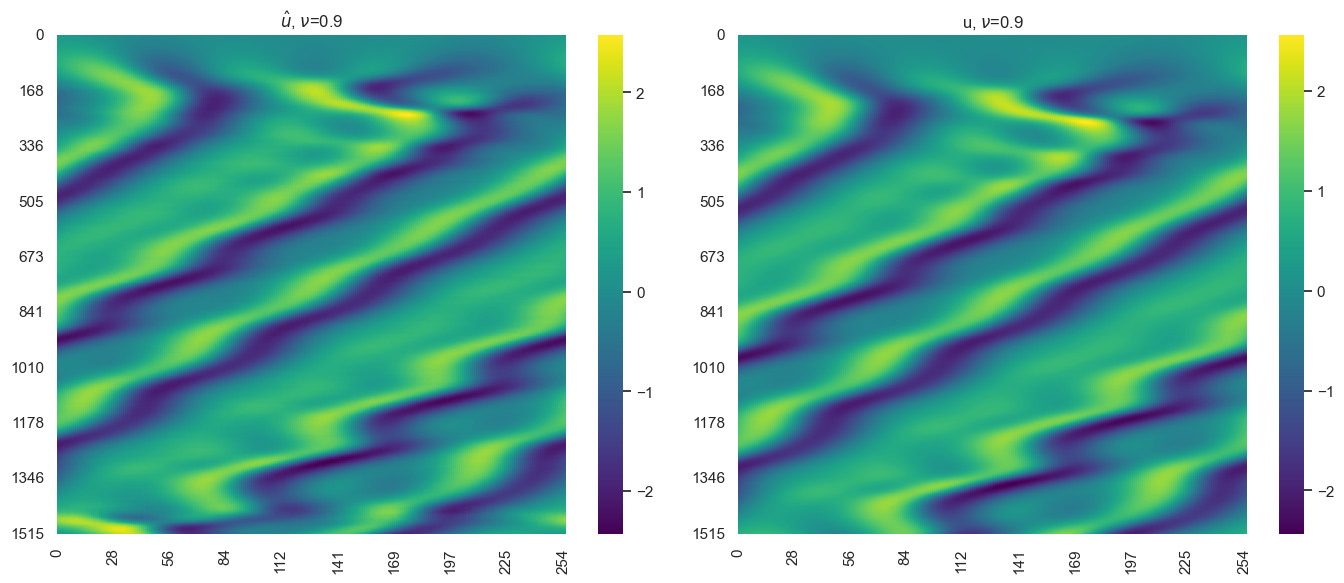} \\
    \includegraphics[width=0.5\textwidth]{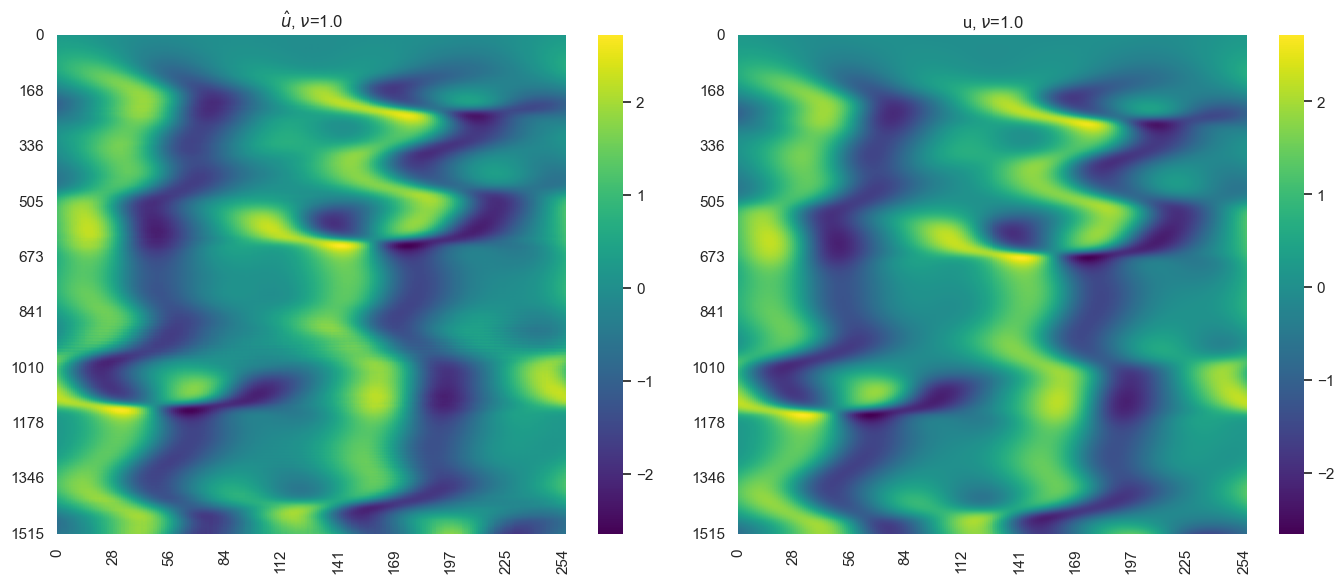} \\
    \includegraphics[width=0.65\textwidth]{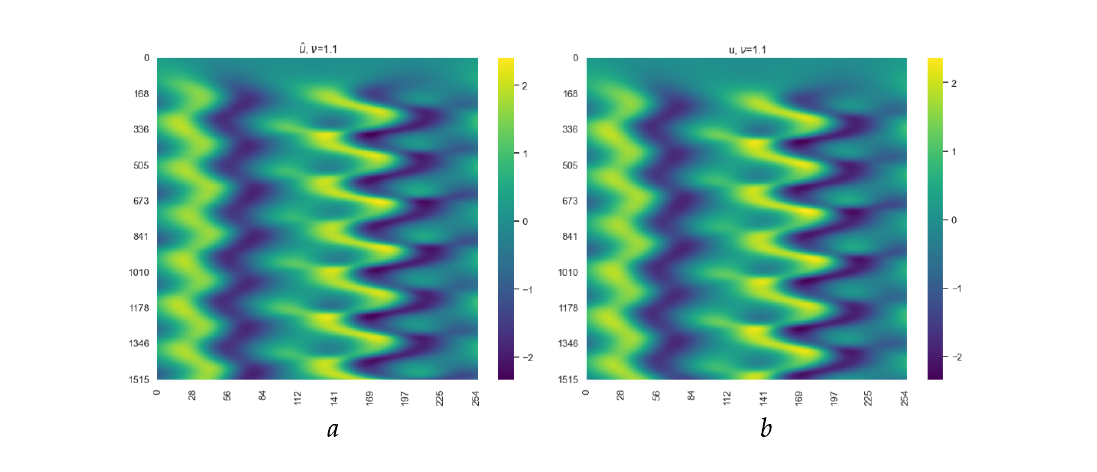}
    \caption{ KS solution \textit{u} field prediction (a) vs true (b) across a range of $\nu$ from 0.7 (top) to 1.1 (bottom)}
    \label{fig:KS}
\end{figure}

We believe that UP-dROM is physics agnostic and could therefore apply to other family of equations, especially the The Kuramoto-Sivashinsky (KS) equation. The KS equation is a nonlinear partial differential equation used as a benchmark in turbulence modelling and the study of chaotic systems. It is given by:

\[
\frac{\partial u}{\partial t} + u \frac{\partial u}{\partial x} + \frac{\partial^2 u}{\partial x^2} + \nu \frac{\partial^4 u}{\partial x^4} = 0
\]

where \( u(x,t) \) represents the field of interest, \( x \) is the spatial coordinate, \( t \) is time, and \( \nu \) is a positive parameter. The KS equation is valued for its ability to capture the essential features of turbulence, such as chaotic behaviour and complex spatiotemporal patterns, while remaining relatively simple and analytically tractable, making it an ideal testbed for developing and validating UP-dROM through turbulent or near turbulent regimes. We compare the inferred and ground truth domains for different values of parameter $\nu$ in Figure \ref{fig:KS}

% Following the same process, it can be observed that the uncertainty increases as $\nu$ goes down, hence towards more chaotic dynamics. 
% \begin{figure}[H]
% \centering
% \includegraphics[width=0.6\textwidth]{KS_UQ.png}
% \caption{ Uncertainty Quantification across the parameter $\nu$ range}
% \end{figure}

\subsection{Model Architecture}

\begin{table}[H]
    \centering
    \begin{tabular}{|l|c|c|c|c|c|} 
        \hline
        & Prediction Horizon & Lookback Window & Space (dim) & Time (dim) & Latent Dimensions \\ 
        \hline
        Navier Stokes & 10 & 10 & $[131\times 100]$ & 3033 & 4 \\ 
        \hline
        KS & 30 & 30 & 255 & 3033 & 32  \\ 
        \hline
    \end{tabular}
    \caption{Model Hyperparameters for Navier Stokes and KS}
\end{table}
\begin{table}[H]
    \centering
    \begin{tabular}{|l|c|c|c|} 
        \hline
        & Hidden Dimensions & Attention Heads & Attention Blocks  \\ 
        \hline
        Navier Stokes & 64 & 8 & 1 \\ 
        \hline
        KS & 128 & 16 & 2  \\ 
        \hline
    \end{tabular}
    \caption{Transformer Hyperparameters for Navier Stokes and KS}
\end{table}
\subsection{Model Loss}
\begin{equation}
    \mathcal{L} = \frac{\lambda}{m} (\sum_{k=0}^{m} \| \phi_{n-q:n} -\mathcal{D}\mathcal{E}({\phi}_{n-q:n}) \| + \beta . \text{KLD}) +  \frac{1}{m} \sum_{k=0}^{m} \| z_{n+1:n+h} - \hat{z}_{n+1:n+h} \| + \frac{1}{m} \sum_{k=0}^{m} \| \phi_{n+1:n+h} -\mathcal{D}(\hat{z}_{n+1:n+h}) \|
\end{equation}

\begin{table}[H]
    \centering
    \begin{tabular}{|l|c|c|} 
        \hline
        & $\lambda$ & $\beta$ \\ 
        \hline
        Navier Stokes & 100 & $1e-4$ \\ 
        \hline
        KS & 100 & $1e-4$   \\ 
        \hline
    \end{tabular}
    \caption{Loss function parameters}
\end{table}
\begin{equation}
    \text{KLD} = -\frac{1}{2} \sum_{i=1}^{Z} \left( \sigma_i^2 + \mu_i^2 - 1 - \log(\sigma_i^2) \right)
\end{equation}

where $\mu$ is the mean vector, $\sigma^2$ is the variance, and $Z$ is the dimensionality of the latent space. 
%\newpage

\end{document}